%% file: root.tex
\documentclass[journal]{IEEEtran}

\input{packages.tex}

\begin{document}

\input{shortcuts.tex}

\input{title.tex}

\maketitle
\input{abstract.tex}
\input{introduction.tex}
\input{related_work.tex}
\input{problem_formulation.tex}
\input{challenges.tex}

\input{approach.tex}
\input{experiments.tex}

\input{conclusions.tex}
\input{appendix.tex}

\bibliographystyle{IEEEtran}
\bibliography{root.bbl}  
\end{document}

%% file: packages.tex
\usepackage[utf8]{inputenc}
\usepackage{amsmath}
\usepackage{amssymb}
\usepackage{graphicx}
\usepackage{cancel}
\usepackage{dsfont}
\usepackage{algpseudocode}
\usepackage{algorithm,algorithmicx}
\usepackage{tikz}
\usepackage{bbm}
\usepackage{dsfont}
\usepackage{graphicx}
\usepackage{amsthm}
\usepackage{dsfont}
\usepackage{graphicx}
\usepackage{subcaption}
\usepackage{multirow}
\usepackage{placeins}
\usepackage{hyperref}

\usepackage{enumerate}
\usepackage{xspace}
\usepackage{romannum}
\usepackage{xcolor,colortbl}
\usepackage{tabularx}
\usepackage[font=scriptsize,labelfont=bf]{caption}

\usepackage{multicol}
\usepackage{wrapfig}
\usepackage{booktabs}

\usepackage{lipsum}

%% file: shortcuts.tex

\newtheorem{theorem}{Theorem}[section]
\newtheorem{assumption}[theorem]{Assumption}
\newtheorem{lemma}[theorem]{Lemma}
\newtheorem{claim}[theorem]{Claim}
\newtheorem{problem}[theorem]{Problem}
\newtheorem{proposition}[theorem]{Proposition}
\newtheorem{corollary}[theorem]{Corollary}
\newtheorem{construction}[theorem]{Construction}
\newtheorem{definition}[theorem]{Definition}
\newtheorem{example}[theorem]{Example}
\newtheorem{xca}[theorem]{Exercise}
\newtheorem{comments}[theorem]{Comments}
\newtheorem{remark}[theorem]{Remark}

\newcommand{\zerob}{\mathbf{0}}
\newcommand{\oneb}{\mathbf{1}}%

\newcommand{\snr}{\text{SNR}}
\global\long\def\norma#1{\left\Vert #1\right\Vert }%

\newcommand{\Ab}{\mathbf{A}}%
\newcommand{\ab}{\mathbf{a}}%

\newcommand{\Bb}{\mathbf{B}}%
\newcommand{\bb}{\mathbf{b}}%

\newcommand{\Cb}{\mathbf{C}}%
\newcommand{\cb}{\mathbf{c}}%

\newcommand{\Db}{\mathbf{D}}%
\newcommand{\db}{\mathbf{d}}%

\newcommand{\Eb}{\mathbf{E}}%
\newcommand{\eb}{\mathbf{e}}%

\newcommand{\Fb}{\mathbf{F}}%
\newcommand{\fb}{\mathbf{f}}%

\newcommand{\Gb}{\mathbf{G}}%
\newcommand{\gb}{\mathbf{g}}%

\newcommand{\Hb}{\mathbf{H}}%
\newcommand{\hb}{\mathbf{h}}%

\newcommand{\Ib}{\mathbf{I}}%
\newcommand{\ib}{\mathbf{i}}%

\newcommand{\Jb}{\mathbf{J}}%
\newcommand{\jb}{\mathbf{j}}%

\newcommand{\Kb}{\mathbf{K}}%
\newcommand{\kb}{\mathbf{k}}%

\newcommand{\Lb}{\mathbf{L}}%
\newcommand{\lb}{\mathbf{l}}%

\newcommand{\Mb}{\mathbf{M}}%
\newcommand{\mb}{\mathbf{m}}%

\newcommand{\Nb}{\mathbf{N}}%
\newcommand{\nb}{\mathbf{n}}%

\newcommand{\Ob}{\mathbf{O}}%
\newcommand{\ob}{\mathbf{o}}%

\newcommand{\Pb}{\mathbf{P}}%
\newcommand{\pb}{\mathbf{p}}%

\newcommand{\Qb}{\mathbf{Q}}%
\newcommand{\qb}{\mathbf{q}}%

\newcommand{\Rb}{\mathbf{R}}%
\newcommand{\rb}{\mathbf{r}}%

\newcommand{\Sb}{\mathbf{S}}%
\renewcommand{\sb}{\mathbf{s}}%

\newcommand{\Tb}{\mathbf{T}}%
\newcommand{\tb}{\mathbf{t}}%

\newcommand{\Ub}{\mathbf{U}}%
\newcommand{\ub}{\mathbf{u}}%

\newcommand{\Vb}{\mathbf{V}}%
\newcommand{\vb}{\mathbf{v}}%

\newcommand{\Wb}{\mathbf{W}}%
\newcommand{\wb}{\mathbf{w}}%

\newcommand{\Xb}{\mathbf{X}}%
\newcommand{\xb}{\mathbf{x}}%

\newcommand{\Yb}{\mathbf{Y}}%
\newcommand{\yb}{\mathbf{y}}%

\newcommand{\Zb}{\mathbf{Z}}%
\newcommand{\zb}{\mathbf{z}}%

\newcommand{\RF}{\mathbb{R}} 

\newcommand{\CF}{\mathbb{C}}

\newcommand{\IF}{\mathbb{I}} 

\newcommand{\NF}{\mathbb{N}}

\newcommand{\define}{\triangleq}%
\newcommand{\const}{\text{const}}%
\newcommand{\argThis}[3]
{
	\underset{#2}{#1}
	\left \lbrace #3 \right \rbrace
}
\newcommand{\argmax}[2]{\argThis{\text{argmax}}{#1}{#2}}
\newcommand{\argmin}[2]{\argThis{\text{argmin}}{#1}{#2}}
\newcommand{\defineInline}[2]{\underset{#2}{\underbrace{#1}}}
\global\long\def\=#1{\underset{#1}{=}}%

\newcommand{\intX}{\intop_{\Omega}}%
\newcommand{\inttheta}{\intop_{\Theta}}%
\newcommand{\intone}{\intop_{-1}^{1}}%
\newcommand{\intinf}{\intop_{-\infty}^{\infty}}%

\newcommand{\tetahat}{\hat{\boldsymbol{\theta}}}%
\global\long\def\cnormaldistribution#1#2{\mathcal{N}^{c}\left(#1,#2\right)}%
\global\long\def\normaldistribution#1#2{\mathcal{N}\left(#1,#2\right)}%
\renewcommand{\tt}{\boldsymbol{\theta}}%
\newcommand{\var}{\text{Var}}%
\newcommand{\E}{\mathbb{E}}%
\newcommand{\pp}{\mathbb{E}}%
\global\long\def\Ehat#1{\hat{\text{\ensuremath{\E}}}\left[#1\right]}%
\newcommand{\cov}{\text{Cov}}%
\global\long\def\pdf#1{f\left(#1\right)}%
\newcommand{\mse}{\text{MSE}}%
\newcommand{\varnoise}{\sigma^{2}}%
\newcommand{\varalpha}{\sigma_{a}^{2}}%
\newcommand{\pr}{\mathbb{P}}%
\newcommand{\independent}{\perp \!\!\! \perp}

\newcommand{\thetaml}{\hat{\theta}_{ML}}%

\global\long\def\diag#1{\text{diag}\left(#1\right)}%
\newcommand{\tr}{\text{tr}}%
\newcommand{\Kroneckerproduct}{\otimes}%
\newcommand{\Hadamardproduct}{\odot}%
\newcommand{\eye}{\mathbf{I}}%

\global\long\def\Real#1{\text{Re}\left(#1\right) }%

\global\long\def\Imag#1{\text{Im}\left(#1\right) }%

\global\long\def\DFT#1{\mathcal{DFT}\left\{  #1\right\}  }%
\global\long\def\IDFT#1{\mathcal{DFT}^{-1}\left\{  #1\right\}  }%
\global\long\def\fft#1{\mathcal{FFT}\left\{  #1\right\}  }%
\global\long\def\ifft#1{\mathcal{FFT}^{-1}\left\{  #1\right\}  }%
\global\long\def\ifourier#1{\mathcal{F}^{-1}\left\{  #1\right\}  }%
\global\long\def\fourier#1{\mathcal{F}\left\{  #1\right\}  }%

\global\long\def\phiphi#1{\frac{\partial}{\partial#1}}

\newcommand{\TV}[1]{#1}		
\newcommand{\VI}[1]{{\color{green}#1}}	
\newcommand{\RM}[1]{{\color{teal}#1}}   	
\newcommand{\CH}[1]{{\color{darkgreen}#1}}	

\newcommand{\xx}{X}%

\global\long\def\nn{\mathcal{N}}%

\newcommand{\cc}{\mathcal{C}}
\newcommand{\ccsize}{|\cc|}

\newcommand{\Aspace}{\mathcal{A}}
\newcommand{\Bspace}{\mathcal{B}}
\newcommand{\Sspace}{\mathcal{S}}
\newcommand{\Xspace}{\mathcal{X}}
\newcommand{\Zspace}{\mathcal{Z}}

\newcommand{\betaalphabet}{\beta_{\alpha,\beta}^k}

\newcommand{\setout}[1][k]{\Omega^{out}_{#1}}
\newcommand{\setin}[1][k]{\Omega^{in}_{#1}}

\newcommand{\xdomain}[1][k]{\Omega_{\xx_{#1}}}

\newcommand{\btildin}[1][k]{\tilde{b}_{#1}^{in}}
\newcommand{\btildout}[1][k]{\tilde{b}_{#1}^{out}}
\newcommand{\Nin}{N_{in}}
\newcommand{\sortedpsi}[1][k]{\tilde{\psi}_{#1}}

\newcommand{\prob}[1]{\ensuremath{\mathbb{P}({#1})}}
\newcommand{\no}{N}
\newcommand{\etab}{\underline{\eta}}
\newcommand{\beliefb}{\underline{b}}
\newcommand{\ee}[1]{\underset{#1}{\mathbb{E}}}
\newcommand{\belieftk}[1][\xx_t]{\pr \left( #1 \mid H_k, a_{k:t-1} \right)}

\newcommand{\xsafe}{\mathcal{X}_{\text{safe}}}
\newcommand{\xunsafe}{\mathcal{X}_{\text{unsafe}}}
\newcommand{\psafe}{\pr_{\text{safe}}}
\newcommand{\punsafe}{\pr_{\text{unsafe}}}

\newcommand{\gdot}{g\left(\cdot\right)}
\newcommand{\belief}[2][k]{b_{#1}\left[#2\right]}
\newcommand{\belieftild}[2][k]{\tilde{b}_{#1}\left[#2\right]}
\newcommand{\geobelief}[2][k]{b^g_{#1}\left[#2\right]}

\newcommand{\Thetasize}{N^{\Theta}}

\newcommand{\PT}{\pr_T}  
\newcommand{\PZ}{\pr_Z}  
\newcommand{\Ivis}{\oneb^{vis}}
\newcommand{\Itime}{I^{time}}
\newcommand{\Iobj}{I^{obj}}

\newcommand{\rectangle}{{%
  \ooalign{$\sqsubset\mkern3mu$\cr$\mkern3mu\sqsupset$\cr}%
}}

\newcommand{\ebfull}{\texttt{theoretical-all-hyp} }
\newcommand{\Ebfull}{\texttt{Theoretical-all-hyp} }
\newcommand{\ebpruned}{\texttt{theoretical-pruned} }
\newcommand{\Ebpruned}{\texttt{Theoretical-pruned} }
\newcommand{\PFfull}{\texttt{PF-all-hyp} }
\newcommand{\PFpruned}{\texttt{PF-pruned} }
\newcommand{\MCMC}{\texttt{MCMC-Ours} }
\newcommand{\SNIS}{\texttt{SNIS-Ours} }
\newcommand{\GSMAP}{\texttt{GS-MAP} }
\newcommand{\NA}{\texttt{ N/A }}

%% file: title.tex
\title{
	Online Hybrid-Belief POMDP with Coupled Semantic-Geometric Models
}


\author{Tuvy Lemberg and Vadim Indelman 
	\thanks{Tuvy Lemberg is with the Department of Aerospace Engineering, Technion - Israel Institute of Technology, Haifa 32000,
		Israel, \texttt{tuvy@campus.technion.ac.il}. Vadim Indelman is with the Department of Aerospace Engineering and with the Department of Data and Decision Science, Technion - Israel Institute of Technology, Haifa 32000, Israel. \texttt{ vadim.indelman@technion.ac.il}. This work was  partially supported by the Israeli Smart Transportation Research Center (ISTRC), US NSF/US-Israel BSF, and the Israel Science Foundation (ISF).
	}
}

%% file: abstract.tex
\begin{abstract}

Robots operating in complex and unknown environments frequently require geometric-semantic representations of the environment to safely perform their tasks. Since the environment is unknown\TV{,} robots must infer \TV{it} and account for many possible scenarios when planning future actions. Because \TV{object} class types are discrete, and the robot's self-pose and the objects' poses are continuous, the environment can be represented by a hybrid discrete-continuous belief\TV{,} which has to be updated according to models and incoming data. Prior probabilities and observation models representing the environment can be learned from data using deep learning algorithms. Such models often couple environmental semantic and geometric properties. As a result, all semantic variables are interconnected, causing the \TV{dimensionality of the semantic state space} to increase exponentially.
In this paper, we consider the framework of planning under uncertainty using partially observable Markov decision processes (POMDPs) with hybrid semantic-geometric beliefs. The models and priors \TV{considered in this work assume} coupling between semantic and geometric variables. We show that obtaining representative samples of the theoretical hybrid belief, required for estimating the value function, is very challenging. As a key contribution, we develop a novel form of the hybrid belief and leverage it to \TV{generate} representative samples.
Furthermore, we show that under certain conditions, the value function and \TV{constraints, such as the probability of safety,} can be calculated efficiently with an explicit expectation over \emph{all} possible semantic mappings.
Our simulations show that our estimators of the objective function achieve similar levels of accuracy compared to estimators that run exhaustively on the entire semantic state-space using samples from the theoretical hybrid belief. Nevertheless, the complexity of our estimators is polynomial rather than exponential. 

\end{abstract}

%% file: introduction.tex
\section{Introduction}


Performing advanced tasks and ensuring safe and reliable operation of autonomous robots frequently requires semantic-geometric mapping of the environment \cite{Velez12jair, chen22rs, Nuchter08ras, Kopitkov18iros, Feldman20arj, Cabello24icra, crespo20as}. In an uncontrolled environment, robot's state, geometric \TV{properties of the} environment such as object locations, and semantic \TV{properties} such as object classes, are often unknown. Therefore, the robot must infer its environment and account for different possible scenarios when planning future actions.

POMDPs \cite{Sondik73or} provide a natural and conceptually abstract framework for robot action planning in the face of uncertainty \cite{spaan04icra, Kurniawati22ar, Kurniawati11rss}. Using POMDPs, the robot maintains a belief over the state, updates the belief based on received observations and actions, and chooses the next actions that maximize the value function while fulfilling the constraints.

Since the robot's state and environment are unknown, they can be represented by the belief's state. The belief, which is a posterior probability of the state, represents the probability of different scenarios. It can be derived using prior probabilities, a motion model, an observation model, and previous history such as previous actions and observations.

\subsection{\TV{Coupled Semantic-Geometric Models}}
Prior probabilities and models can be \TV{learned} using machine learning and deep learning algorithms. \TV{The learned priors can account for the dependencies between geometric and semantic variables. For instance, in \cite{cherabier19eccv}, a semantic 3D reconstruction  methodology is proposed, where the model learns priors that capture the dependencies between semantic labels and the geometric properties of the environment. By doing so, they achieve improved accuracy in 3D reconstruction tasks, while reducing the number of parameter of the neural network. Likewise, learned prior probabilities can link an object's class and its location, since certain objects are more likely to be found in certain locations \cite{gervet23robotics}.}

In general, semantic observations often depend on both the class of the observed object and its relative position to the robot \cite{Kopitkov18iros, Feldman20arj, Tchuiev23ai, Cabello24icra, Yang19tro}.  For example, an image-based classifier may produce different results for the same object when the images are taken from different viewpoints.

\TV{
Manhardt et al.\ \cite{manhardt19iccv} study the ambiguity of object detection and 6D pose estimation from visual data. They highlight that objects often appear similar from different viewpoints due to shape symmetry, occlusion and repetitive textures.
A bowl and a mug can appear similar from certain poses, leading to potential misclassification and incorrect pose estimation errors. A robot tasked with locating the mug should account for this ambiguity in its planning. 
}

\subsection{\TV{Hybrid Semantic-Geometric Belief in SLAM and POMDPs}}
The coupling between the object's pose and its class causes statistical dependency between all random variables of the state. Therefore, the classes of all objects are mutually dependent. Since all classes are dependent, the number of semantic mapping hypotheses is exponential in the number of objects. This makes many required computations, such as finding the most probable hypothesis and computing the value function, computationally prohibitive.

Many POMDP solvers approximate the value function using sample-based estimators \cite{Sunberg18icaps, Somani13nips, Silver10nips, Lim23jair}. This is a common approach in POMDPs, since computing expectations over continuous state space is often intractable, the state space dimensionality can be substantial and the horizon the robot must plan for can be far, making numerical approximation difficult.   
However, this approximation introduces estimation error, which can be substantial in the case of hybrid semantic-geometric beliefs since the state space increases exponentially with the number of objects.

Only a few studies considered a semantic-geometric belief with dependency between classes and poses, and most of them are in the context of semantic simultaneous localization and mapping (SLAM). Most of these studies use only the maximum likelihood of a classifier output, without considering the coupling between the semantic observation and the relative viewpoint \cite{Pillai15rss, Yang19tro, Moreno13cvpr}. These approaches do not consider a belief over the semantic mappings and therefore cannot assess uncertainty.

The approach proposed in \cite{Doherty20icra}, maintains a belief on classes of objects. However, semantic observations are considered independent of the relative position of the object to the robot. Even though this observation model simplifies the problem computationally, it neglects the effect of viewpoint on classification and vice versa, losing important information. The works \cite{Ginting23arxiv, Ginting24arxiv} use separate beliefs for the robot state and the object's object property.

Recent studies have used viewpoint dependent semantic observation models in SLAM and considered the coupling between object's class and pose, showing improved data association (DA), localization, and semantic mapping \cite{Feldman18icra,Tchuiev19iros,Kopitkov18iros, Feldman20arj, Tchuiev23ai}. However, these works manage the computational burden by pruning most semantic hypotheses. Aside from the performance loss, pruning hypotheses will make it impossible to assess the system's risk if one of the pruned hypotheses turns out to be true. 
Moreover, after pruning and renormalization, the resulting belief is overconfident relative to the original belief \cite{Lemberg22iros}. This may lead to actions that pose a higher risk.

In \cite{Lemberg22iros}, a method was developed 
for estimating the normalization factor and the probability of a single semantic mapping hypothesis considering all possible semantic mappings. Computation of the theoretical normalization factor requires summation over all possible semantic mappings and integration over the continuous state which may be intractable. Although the number of possible semantic mappings is exponential, it was shown that an explicit summation over these mappings for the normalization factor can be calculated very efficiently, without explicitly running through all possible semantic mappings explicitly. The estimation of the normalization factor allows to estimate the probability of a single semantic mapping hypothesis, without running through all possible semantic mappings.

\subsection{\TV{Contributions and Paper Organization}}
In this work, we consider a semantic-geometric hybrid belief POMDP. Observation models and prior probabilities that link between an object's class and its pose, cause all classes to be mutually dependent. This makes the state space increase exponentially with the number of objects. Large space spaces, also known as the curse of dimensionality, is one of the most discussed topics in POMDPs \cite{Kurniawati22ar}.  

Because the number of semantic mapping hypotheses is exponential, explicitly marginalizing the unnormalized belief over the semantic mapping hypotheses is computationally prohibitive. Yet, we show that this can be accomplished efficiently using novel formulations of the belief, allowing us to calculate the marginal unnormalized belief of the continuous state while accounting for \emph{all} the semantic mapping hypotheses. 

Using the marginal unnormalized belief of the continuous state, we show that sampling the belief at planning time can be done efficiently. This is possible by first drawing representative samples of the continuous state and then sampling objects classes given the continuous state samples. Since the models considered coupling between an object's class and its pose, classes of different objects are independent given objects' poses. This allows us to sample semantic mapping hypotheses efficiently.   

Continuous state samples can be obtained using MCMC methods such as the Metropolis-Hastings (MH) algorithm \cite{Beichl00cse} or proposal distribution for self-normalized importance sampling (SNIS) \cite{Torr03pami}. Both require evaluating the marginal unnormalized belief for sampled continuous state realization. This calculation can be performed efficiently using our formulation. In contrast, without it, this evaluation is computationally intractable due to the exponential number of semantic mapping hypotheses. 

Moreover, we demonstrate that using importance sampling to estimate the value function in hybrid POMDP settings can lead to an exponential increase in the mean squared error (MSE) with the number of objects, due to semantic samples. We show that in many cases, the MSE caused by semantic samples has a lower bound that increases exponentially with the number of objects. However, by utilizing our sampling methods, this lower bound is zero in these cases. Furthermore, our empirical simulations demonstrate that the error does not increase with the number of objects. 

To improve estimation error even further, we prove that the objective function can be estimated with an \emph{explicit} expectation over all possible semantic mapping hypotheses efficiently. This can be applied in an open loop setting, where policies are reduced to a pre-defined action sequence, and for a specific structure of reward functions. Since we run through all semantic hypotheses, the true hypothesis will be considered. In contrast, using pruning or sampling hypotheses we may miss the true semantic hypothesis. Furthermore, the Rao-Blackwell theorem \cite{Blackwell47ams} guarantees a reduced estimation error. 

\TV{The probability of safety \cite{Santana16aaai,Moss24arxiv, Carr23aaai,Zhitnikov22ai} in the context of hybrid semantic POMDP is used as an illustrative example for the structured reward throughout this study.}
Often, robots operating in complex environments should satisfy constraints dependent on both geometric and semantic properties. For instance, autonomous vehicles should assess traffic sign type and location.
\TV{In a POMDP, computing the probability of fulfilling such constraints in the future is often called the probability of safety.}

Since each object creates a constraint on the robot's state, and each semantic mapping hypothesis implies a set of constraints, to estimate the probability of safety, one must run through all possible semantic mappings and compute the probability of fulfilling the constraints under each hypothesis. Nevertheless, we show that the probability of safety falls under the special reward function type and therefore explicit expectation of semantic hypotheses can be applied effectively.

To summarize, our main contributions are as follows:
\begin{itemize}
	\item 
	We present a novel formulation for the hybrid semantic-geometric belief. This formulation allows us to calculate the unnormalized marginal belief of the continuous state efficiently, marginalizing over all semantic mapping hypotheses.
	\item 
	Sampling from a belief at planning time is generally intractable. We provide sampling methods that utilize the unnormalized marginal belief of the continues state.
	\item
	Under simplified assumptions we show that the objective function and the probability of safety considering hybrid semantic-geometric beliefs can be estimated efficiently, while running through all possible semantic mappings, thus reducing estimation error.
\end{itemize}

The paper is organized as follows. 
Section \ref{sec: related work} provides an overview of related works. 
In Section \ref{Sec:problem_formulation}, a semantic-geometric belief is formulated followed by POMDPs derivation using the semantic-geometric belief. 
Next, in Section \ref{sec: challenges}, the main challenges associated with this hybrid POMDPs are discussed. 
In Section \ref{sec:approach}, we introduce our proposed methodology. 
Section \ref{Sec:experiment} details the synthetic experiments results, 
and conclusions are discussed in Section \ref{Sec:conclusions}.

%% file: related_work.tex
\section{Related Work} \label{sec: related work}

\subsection{\TV{Semantic SLAM and Viewpoint-Dependent Models}}

Only a \TV{few} semantic SLAM works \TV{have} considered a viewpoint dependent semantic observation model. In such models, observations depend \TV{on the object's class, the object's pose, and the robot's pose.} There are many works outside of SLAM \TV{framework} that estimate the pose of an object depending on its class from a single image, such as \cite{rogez19pami, kolotouros19cvpr, grabner18cvpr}, \TV{thereby} modeling the \TV{coupling}. \TV{However,} in a SLAM framework\TV{,} this coupling causes all classes to be dependent, \TV{therefore} it is very common to neglect \TV{this coupling and consider classes and poses separately.}

Segal and Reid \cite{Segal14iros} implemented a hybrid continuous-discrete belief optimization method for robot localization and mapping. They \TV{approximated} inference on a factor graph with junction nodes.  
Velez et al.\ \cite{Velez12jair} and Teacy et al.\ \cite{Teacy15aamas} utilized an online planning algorithm that considers a viewpoint-dependent semantic observation model \TV{and learns} the spatial correlations between observations. However, they \TV{simplified} the belief by assuming known localization and no prior information is given\TV{.} Therefore, classes of different objects are independent.

Bowman et al.\ \cite{Bowman17icra} formulated the full joint belief over the environment's semantic and geometric properties. They used the expectation maximization \TV{(EM) method} to infer the belief. They showed results in scenarios with a small number of objects and possible classes. Because the number of semantic hypotheses increases exponentially, it is not feasible to \TV{evaluate} all hypotheses in EM.

Feldman and Indelman \cite{Feldman20arj} developed a method to extend a single image classifier into a viewpoint\TV{-}dependent observation model that also captures the model\TV{'s} epistemic uncertainty \cite{Gal16icml}, which \TV{occurs when the} classifier receives raw data at deployment that is far from the data the classifier was trained on. They model the output as a Gaussian process. They show empirically that their method improve\TV{s} robustness and classification accuracy compared to other methods that \TV{do} not use a viewpoint\TV{-}dependent semantic model. Tchuiev and Indelman \cite{Tchuiev23ai} fused the viewpoint-dependent semantic model with epistemic uncertainty in belief space planning.

Morilla-Cabello et al.\ \cite{Cabello24icra} proposed a generalization of the viewpoint\TV{-}dependent observation model. The\TV{ir} observation model \TV{additionally} account\TV{s} for environmental effects, such as appearance, occlusions, and backlighting. They showed improvements in state estimation quality.

Tchuiev et al.\ \cite{Tchuiev19iros} showed that a semantic viewpoint\TV{-}dependent observation model can be used to enhance localization\TV{,} DA\TV{,} and semantic mapping in ambiguous scenarios. To reduce the computational burden, they considered only a few semantic mapping hypotheses and pruned the rest. Kopitkov and Indelman \cite{Kopitkov18iros} used a neural network to learn a viewpoint-dependent measurement model of CNN classifier output features.

\subsection{\TV{Hybrid Discrete-Continuous POMDPs}}

General state-of-the-art POMDP algorithms such as POMCPOW, \TV{PFT-DPW, and} DESPOT \cite{Sunberg18icaps,Somani13nips, Silver10nips} are not explicitly formulated for hybrid discrete-continuous beliefs but can be modified to support \TV{them}. However, these methods assume that the state can be sampled from the belief at planning time. This is not possible in our setting since it requires full knowledge of the probability of semantic hypotheses. Because the number of semantic hypotheses is exponential, it is not feasible to calculate the probabilities of all hypotheses.

Recently, POMDPs with hybrid continuous and discrete states have been investigated in the context of DA \cite{Pathak18ijrr, Shienman22isrr, Barenboim23ral}. 
Several works have incorporated discrete DA variables into the state, creating a hybrid continuous-discrete belief in POMDP. Although the formulations of beliefs with DA are similar to the hybrid belief with semantic-geometric coupling, there are several fundamental differences. 
In DA POMDP, the discrete state space grows exponentially with history. In our case, the discrete state space grows exponentially with the number of objects; thus, it is exponentially large at planning time, but it does not increase within future time instances within planning\TV{,} assuming one does not model observations of new objects that are unknown at planning time.

Barenboim et al.\ \cite{Barenboim23ral} utilized the Monte Carlo Tree Search (MCTS) algorithm to support a hybrid discrete-continuous state POMDP. They propose a sequential importance sampling method with resampling to estimate the value function. They define the proposal distribution for the discrete state to be uniform. In our case, this proposal will result in an exponential MSE. Additionally, we provide several methods for obtaining representative samples that result in significantly lower MSEs.

\subsection{\TV{Rao-Blackwellization in Hybrid Models}}

Several works used the Rao-Blackwell theorem \cite{Blackwell47ams} to reduce the estimation error. 
Doucet et al.\ \cite{Doucet00a} introduced the Rao-Blackwellised Particle Filtering (RBPF). They show that \TV{for some cases,} it is possible to exploit the structure of a Bayesian network by sampling only the state variables that cannot be marginalized analytically\TV{,} and \TV{marginalize} analytically the rest. Using this approach, they improve the particle filter algorithm in two ways: they improve state estimation according to the Rao-Blackwell theorem, and they avoid the curse of dimensionality by reducing sample state dimensionality.

In a recent paper \cite{Lee24arxiv}, the RBPF was integrated with the POMCPOW algorithm \cite{Sunberg18icaps} combined with quadrature-based integration, showing fewer samples were required, and planning quality superiority compared to other methods.

In our work, we utilize the Rao-Blackwell theorem by calculating the expectation over the semantic mapping hypotheses explicitly. This allows us to reduce the MSE of the estimated objective function. We show that this is possible for specific reward structures and open-loop policies. It can be applied to a variety of objectives, such as object search and probability of safety. 
We provide a complementary proof of the Rao-Blackwell theorem for our setting\TV{,} where part of the expectation is estimated using samples and the other part is computed explicitly (see Theorem \ref{theorem: Rao-Blackwell for samples}).

%% file: problem_formulation.tex
\section{Preliminaries and Problem Formulation \label{Sec:problem_formulation}}

In this section, we formulate the agent's posterior probability distribution, also known as belief, where a semantic, viewpoint-dependent observation model is considered. This belief is hybrid, containing both continuous-geometric variables, such as agent's and objects' poses, and discrete-semantic variables, such as objects' classes. Next, we formulate the hybrid semantic-geometric POMDP. Lastly, we introduce semantic risk awareness under \TV{a} semantic-geometric POMDP with semantic-geometric constraints. 

\subsection{Hybrid Semantic-Geometric Belief}
\label{subsec:belief}

Consider an agent operating in an unknown environment. During operation, 
the agent maps the environment and classifies objects to perform its tasks.
Since the environment is unknown and only partially observable, 
it is often represented using random variables and statistical models that represent the relations between them. 
At time-step $k$, define the agent's state as $x_k$, and the $n$th
object's state and class as $x^o_n$ and $c_n$, respectively. Each object is assumed to 
belong to a single class out of $N^c$ classes, thus
$c_n \in \left[1,\ldots,N^c\right]$, for the $n$th object. 
We assume that the object's state and class do not change 
over time.

 Define the number of objects encountered by the agent up to time-step $k$, as $N^o_k$. The subscript $k$ will be omitted to simplify the notation while always considering $N^o$ at planning time $k$. In addition, we define $x_{1:k}=\left\{x_1, \ldots, x_k\right\}$ as the agent's trajectory from time-step $1$ up until time-step $k$, and $X^o=\left\{x^o_1, \ldots x^o_{N^o}\right\}$ as the concatenation of the objects' states. 

The concatenation of all unknown continuous variables is defined as 
$\xx_k\define\left\{ X^o,x_{1:k}\right\} $ with a corresponding continuous state space $\Xspace$.
A hypothesis is defined as the concatenation of all objects' classes, also called semantic mapping,
$
C\define\left\{ c_{1},\ldots,c_{N^{o}}\right\} 
$,
with the domain
$\cc$. Since $\cc$ consists of all possible semantic mappings,
its size is $|\cc|=(N^c)^{N^o}$.

The hybrid semantic-geometric belief at time-step $k$ is defined as follows
\begin{equation}
	b_{k}\left[\xx_k,C\right]\define\pr\left(\xx_k,C\mid H_k\right),
	\label{eq:belief}
\end{equation}
where $H_k \define \left\{ a_{0:k-1},z_{1:k}\right\}$ is the history, $a_{0:k-1}$ is the action sequence from time-step $0$ up until time-step $k-1$, and $z_{1:k}$ is the observation sequence from time-step $1$ up until time-step $k$.
This structure leverages geometric information contained in both the semantic and geometric observations, as defined below, rather than assuming geometric and semantic variables to be independent. However, this comes at a cost of increased complexity.
Finally, we define the \TV{belief's state} as $S_k \define (\xx_k, C)$, with the state space $\Sspace = \Xspace \times \cc$.

The belief \eqref{eq:belief} can be written recursively using Bayes' theorem followed by the chain rule.
\begin{multline}
	b_k\left[\xx_k,C\right]
	= \\ 
	\tilde{\eta}_k
	\PZ\left(z_k\mid X^o, x_k,C\right)
	\PT\left(x_k \mid a_{k-1}, x_{k-1}\right)
	b_{k-1}\left[\xx_{k-1},C\right],
	\label{eq:belief recursive}
\end{multline}
where $ \tilde{\eta}_k = \left(\pr\left(z_k \mid H_{k-1}, a_{k-1} \right)\right)^{-1}$ is the normalization factor,
the transition model,
$\PT\left(x_{k} \mid x_{k-1},a_{k-1}\right)$,
describes the probability of moving from state $x_{k-1}$ to state $x_{k}$ by taking action $a_{k-1}$, 
and the observation model 
$\PZ\left(z_k\mid X^o, x_k,C\right)$ 
will be defined in \eqref{eq: observations}.
Following \eqref{eq:belief recursive}, the marginal $\belief{C}$ and conditional $\belief{\xx_k \mid C}$ recursive formulation is given by
\begin{multline}
	b_k\left[\xx_k \mid C\right]
	= \\
	\tilde{\eta}_k^C
	\PZ\left(z_k\mid X^o, x_k,C\right)
	\PT\left(x_k \mid a_k, x_{k-1}\right)
	b_{k-1}\left[\xx_{k-1} \mid C \right],
	\label{eq:belief component recursive}
\end{multline}
where $\tilde{\eta}_k^C = \left(\pr\left(z_k \mid H_{k-1}, a_{k-1}, C \right)\right)^{-1}$, and 
\begin{equation}
	b_k\left[C\right]
	= 
	\frac{
		\tilde{\eta}_k
		}
		{
		\tilde{\eta}_k^C
		}
	b_{k-1}\left[C\right].
	\label{eq:belief weight recursive}
\end{equation}
Since objects are not always visible, the set of visible objects at time step $t$ is defined as 
$
\Iobj[t] 
$ 
for $t \in \left[1,\ldots,k\right]$.
Given DA, this set is deterministic and known at planning time.
Generally, DA is unknown and should be included in the belief's state \cite{Tchuiev19iros}. 
However, to simplify the analysis, DA is assumed to be known in this study.

\begin{figure}[H]
	\centering
	\includegraphics[width=0.75\linewidth]{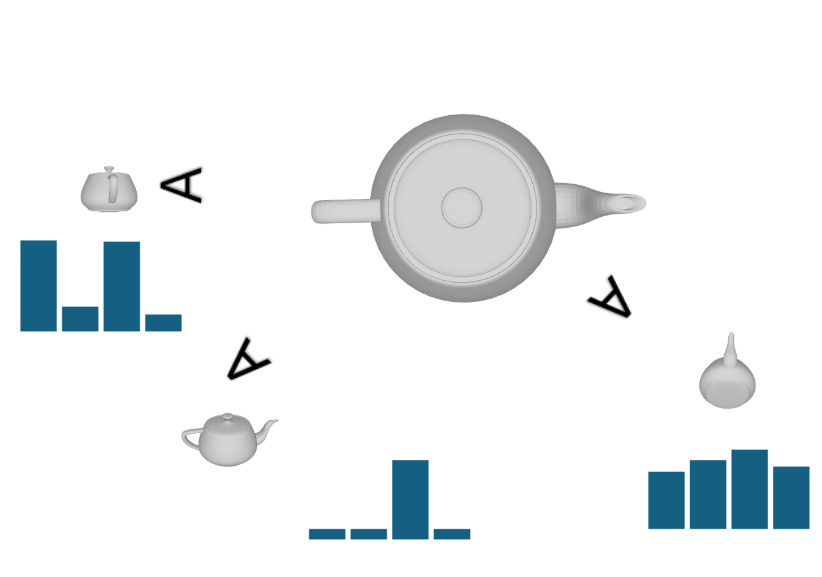}
	\caption{An illustration of a viewpoint-dependent observation model. Sensors receive different information from different viewpoints. Consider the semantic observations are the output of a classifier. The blue bars represent classifier output. From each viewpoint the semantic observation, in this case the distribution over classes obtained from a classifier, can vary due to change in visual appearance or detected features.}
	
	\label{fig:IllustrationViewpointDependent}
\end{figure}

At time step $k$, $ z_k $ consists of observations of visible objects, 
$z_k=\left\{z_{k,n}\right\}_{n\in\Iobj[k]}$.
Consequently, the observation model is given by 
\begin{equation}
\PZ\left(z_k \mid X^o,x_k, C\right)
=
\prod_{n \in \Iobj[k]}
\PZ\left(z_{k,n} \mid x^o_n, x_k, c_n \right),
\label{eq: observations}
\end{equation}
Observation $z_{k,n}$ consists of a semantic part $z^s_{k,n}$.
\TV{
The semantic observation $z^s_{k,n}$ depends on the object's class $c_n$, and may also depend on the robot's and object's poses. This definition supports a viewpoint-dependent semantic observation model.  $z^s_{k,n}$ can represent the output of a classifier, for example.
Consider a semantic viewpoint dependent observation model for example. Viewpoints from which different classes are visibly similar are taken into account. This can be very useful in planning tasks where the agent needs to disambiguate between different classes. This can be very useful in planning tasks where the agent needs to disambiguate between different classes.
In contrast, the geometric observation $z^g_{k,n}$ is independent of the object's class and can represent various types of measurements, including, for example, visual landmarks, LIDAR and  GPS measurements.
}
Given $x^o_n$, $x_k$, \TV{and $c_n$}, the observations $z^s_{k,n}$ and $z^g_{k,n}$ are independent, thus
\begin{multline}
	\PZ\left(z_{k,n}\mid x_k,x_n^o,c_n\right)
	=\\
	\PZ\left(z_{k,n}^s\mid x_k,x_n^o,c_n\right)
	\PZ\left(z_{k,n}^g\mid x_k,x_n^o\right).
	\label{eq: observation model}
\end{multline}
Figure \ref{fig:IllustrationViewpointDependent} illustrates a viewpoint-dependent observation model.

Two reasons cause coupling between continuous and discrete variables. 
First, the semantic observation model involves 
$x_k, x^o_n$ and $c_n$, resulting in their coupling. 
Secondly, $\xx_0$ and $C$ can be dependent in the prior belief. 
Additionally, given $\xx_0$ the classes of different objects are assumed independent, thus
\begin{align}
	\pr_0\left(\xx_0, C\right)
	&=
	\pr_0\left(\xx_0\right)\pr_0\left(C \mid \xx_0 \right) \nonumber
	\\ &= 
	\pr_0\left(\xx_0\right)
	\prod_{n=1}^{N^o}  \pr_0\left(c_n \mid \xx_0\right).
	\label{eq:PriorProbability}
\end{align}
\TV{
\textit{Intuition and motivation.}
In realistic settings, $c_n$ and $\xx_0$ are dependent. The conditional prior \eqref{eq:PriorProbability} encodes contextual knowledge that certain classes are more (or less) likely to be found at certain locations. For example, consider a robot that operates in an apartment and has prior information about the apartment's room types and locations. It is possible to obtain such information using  Kimera \cite{Rosinol21ijrr}, for example. 
Consider the robot being asked to make a list of groceries for ordering. It should search for groceries in the kitchen, since this is the most likely location to find them. 

In general, object classes may be linked a priori. For instance, given that a keyboard is located at $x^o_1$, it is likely that a mouse is nearby. Later, we will see that, by using the prior \eqref{eq:PriorProbability}, the posterior belief \eqref{eq: belief X and C|X} maintains the same structure of conditional independence of the classes given the states, enabling efficient computation for  planning.	
}

\subsection{POMDP with Hybrid Semantic-Geometric Belief \label{subsec:POMDP}}

Considering a planning session at time instant $k$.
A finite horizon $\rho$-POMDP can be defined as the tuple
$\left\langle \Sspace,\Aspace,\Zspace,\PT,\PZ,\rho, b_k\right\rangle $,
where $\Sspace, \Aspace,\Zspace$ are the state, action and  observation spaces, respectively; 
$\PT,\PZ$, are the transition and observation models, 
$\rho$ is the reward function;
and $b_k$ is the hybrid posterior belief, previously defined in \eqref{eq:belief}.
$L$ is the planning horizon.
$\PT,\PZ$ and $\Sspace$ are the same as previously defined in section \ref{Sec:problem_formulation}.
The observation space $\Zspace$ consists of the semantic and geometric observations spaces respectively, $\Zspace=\Zspace^s \times \Zspace^g$.  

A policy at time-step $t$ is a function from the belief space $\mathcal{B}$ to the action space $\Aspace$, $\pi_t: \mathcal{B}\to \Aspace$.
Given a sequence of policies $\pi_{k:L}$ and the POMDP tuple, the value function is defined as follows
\begin{equation}
	V_k\left(b_k, \pi_{k:L}  \right)
	=
	\ee{z_{k+1:L}}
	\left[
		\sum_{t=k}^L 
		\rho_t \left( b_t, \pi_t(b_t) \right) \mid b_{k}, \pi_{k:L}
	\right]
	.
	\label{eq:value function}
\end{equation}
The value function can be formulated recursively as
\begin{equation}
	V_k\left(b_k, \pi_{k:L}  \right)
	=
	\rho_k \left( b_k, \pi_k(b_k) \right)
	+ \ee{z_{k+1}}
	\left[
		V_{k+1}\left(b_{k+1},\pi_{k+1:L}\right)
	\right]
	.
	\label{eq:value function recursive}
\end{equation}	
The objective of the POMDP is to find the optimal policies that maximize the value function,
$
\pi_{k:L}^{\star}
=
\argmax{\pi_{k:L}}{V_k\left(b_k, \pi_{k:L} \right)}
$.
Typically, the reward is defined as the expectation of function $r_t$, that \TV{can be} a function of the belief, the state, and the action $a_t=\pi_t(b_t)$,
\begin{equation}
\rho_t \left(b_t, a_t\right)
=
\ee{\belief[t]{\xx_t,C}}
\left[
r_t \left(a_t, C, \xx_t, \belief[t]{C, \xx_t} \right)
\right]
.
\label{eq:reward}
\end{equation}
$r_t$ is named the inner reward function.
\TV{
Since the inner reward is time dependent, it may include a discount factor 
to account for increasing uncertainty in future predictions, thereby balancing immediate and future rewards. Thus $\rho_t(b_t, a_t) = \gamma^t \E [ r_t(\cdot)] =   \E [ \gamma^t r_t(\cdot)]$ where $\gamma \in (0,1]$.
}
Furthermore, $\rho_t$ supports state-dependent rewards $r_t = r_t(a_t, C, \xx_t)$
and information-theoretic rewards $r_t = r_t(a_t, \belief[t]{C,\xx_t})$ such as entropy, in which case $r_t = -\log(b_t[C,X_t])$.

\TV{
In this work, we define open-loop and closed-loop settings as follows: 
Closed-loop setting refers to general policies $\pi_{k:L}$ where the policy 
$\pi_t$ is a mapping from the belief space $\mathcal{B}$ to the action space $\Aspace$. Thus, it is dependent on future observations $z_{k+1:t}$.
In contrast, open-loop setting refers to  predefined a set of action sequences, where each action sequence  $a_{k:L}$  remains constant regardless of future observations.
}

\begin{lemma}
	\label{lemma: expected reward action sequence}
	Let $\rho_t$ be defined in \eqref{eq:reward}.
	The value function \eqref{eq:value function} can then be formulated as follows
	\begin{multline}
		\label{eq: value function general reformulated}
		V_k\left(b_k, \pi_{k:L} \right) 
		=\\
		\ee{b_k \left[C, \xx_k\right]}
		\left[
		\prod_{\tau=k+1}^L
		\ee{ \PT\left(x_{\tau} \mid \pi_{\tau-1},x_{\tau-1} \right)}
		\ee{\PZ\left(z_{\tau} \mid \xx_{\tau},C\right)}
		\left[
		\sum_{t=k}^{L}
		r_t\left( \cdot \right)
		\right]
		\right],
	\end{multline}
	where to reduce clutter we use $\pi_{\tau} = \pi_{\tau}\left(b_{\tau}\right)$.
	Further, for an open-loop setting, $\pi_{k:L} \equiv a_{k:L}$, and  state-dependent rewards, the corresponding objective function $J_k$ can be simplified as 
	\begin{multline}
	J_k\left(b_k, a_{k:L} \right)=
	\\
	\ee{b_k \left[C, \xx_k \right]}
	\left[
	\prod_{\tau=k+1}^L
	\ee{\PT\left(x_{\tau} \mid a_{\tau-1},x_{\tau-1}\right)}
	\left[
		\sum_{t=k}^{L}
		r_t\left(a_t, C,\xx_t \right)
	\right]
	\right]
	.
	\label{eq: objective function state dep reward}
	\end{multline}
\end{lemma}
The proof of Lemma \ref{lemma: expected reward action sequence} is provided in Appendix \ref{sec: proof of lemma expected reward action sequence}.

\TV{
For clarity, expectations are expressed either with the explicit distribution of the expectations, as in \eqref{eq:reward}, or with the variable of the expectation, as in \eqref{eq:value function} and \eqref{eq:value function recursive}. 
The majority of the derivations support both open-loop setting and closed-loop setting.
If only open-loop settings are supported, it will be mentioned and $a_{k:L}$ will be used instead of $\pi_{k:L}$.
}

\subsection{Semantic Risk Awareness} 
\label{subsec:semantic safety}

There are many studies that focus on chance constraints and risk awareness within POMDP settings, such as \cite{Santana16aaai, Moss24arxiv, Carr23aaai}. 
In our work, we formulate the probability of safety in an object-semantic-geometric POMDP setting, where both the semantic and geometric properties of the environment affect \TV{the} constraints that the agent must satisfy. This scenario is very common. For example, self-driving cars that encounter traffic signs will have different constraints for each traffic sign; a robot that navigates a cluttered household may have to avoid fragile objects of a certain type while engaging with other types. In both cases, the object type affects the constraints.

To our knowledge, this work is the first to formulate the probability of safety in object-semantic-geometric POMDP settings. We consider the safety constraints to be class dependent. Further, we \TV{show} in this section that computing the probability of safety \TV{is} computationally intractable \TV{if done} naively in a brute-force manner, whereas in Section \ref{sec: structured reward} we show that \TV{using our methods, the probability of safety} can be computed with great efficiency, reaching real-time performance.

Consider scenarios where the agent poses a risk to the environment or vice versa. This event \TV{is} defined by the agent's state, the objects' states, and the objects' classes, \TV{creating constraints on the robot}. 
Because a POMDP setting is considered, it is not always possible to guarantee the fulfilment of the conditions. Instead, the probability of fulfilling the constraints should be considered. In particular, each semantic mapping defines a different constraint on the agent's state, and to compute the probability of safety, it is necessary to run over all possible semantic mappings. Generally, calculating this probability is prohibitively expensive.

Specifically, \TV{suppose each object defines an unsafe subspace for the agent,}  
$\xunsafe \left(c_n, x^o_n\right) \subseteq \mathcal{X}, \;\; n=1,\ldots,N^o$.
Given a semantic map $(C, X^o)$, the unsafe subspace is the union of all objects' unsafe subspaces,
$\xunsafe \left(C, X^o\right) =
\cup_{n=1}^{N^o} \xunsafe \left(c_n, x_n^o\right)$.

\begin{definition}
The probability of safety $\psafe$ is defined as the probability that the agent will only pass through the safe subspace. This is equal to the probability that it will never pass through the unsafe subspace,
\begin{equation}
\psafe
=
\pr \left(
	\left\{
		\wedge_{\tau=k+1}^L x_{\tau} \notin \xunsafe\left(C, X^o\right)
	\right\}
	\mid
	b_k, \pi_{k:L-1}  
\right).
\label{eq:psafe}
\end{equation}
\end{definition}
\vspace{5pt}
In Section \ref{sub: structured reward psafe}, equation \eqref{eq: safety inner reward}, we show \TV{that the probability of safety,} $\psafe$, can be formulated as an expectation of a state-dependent reward, similar to the value function in \eqref{eq: value function general reformulated in proof}.

Therefore, $\psafe$ can be computed using the same methods we developed for the value function. 
Ultimately, we will show that $\psafe$ has a special structure, \TV{as given in} \eqref{eq: efficient reward}, 
allowing us to compute it \TV{efficiently while} considering all possible \TV{combinations of} semantic mappings, assuming open-loop settings.

%% file: challenges.tex
\section{Challenges}
\label{sec: challenges}

Since $C$ is a discrete variable with a finite domain, the expectation of $C$ can principally be computed exhaustively.
However, since $|\cc|$ \TV{grows exponentially with the} number of objects $N^o$, computing \TV{the} explicit expectation over $C$ is impractical. \TV{Additionally}, the expectation over $\xx_k$ does not have an analytical solution in the general case and is therefore intractable. Therefore, sampling-based POMDP solvers \TV{are used to estimate} the value function e.g., \cite{Sunberg18icaps, Somani13nips, Silver10nips, Lim23jair}.

In this setting, one of the main challenges is to obtain samples directly from the belief $b_k$ or from a proposal distribution for importance sampling. 
The proposal distribution \TV{must} be carefully chosen. Otherwise, the estimation error can be extremely high, requiring a significant number of samples.
In the following section we discuss the challenges associated with obtaining samples from the belief at planning time $\belief{\xx_k,C}$. 
Next, we discuss the challenges associated with computing an explicit expectation over $C$.

\subsection{Sampling the Posterior Belief at Planning Time}
\label{subsec: sampling belief}

Consider estimating the value function with samples.
According to equation \eqref{eq: value function general reformulated}, samples of the current state 
$\xx_k^{(i)}, C^{(i)} \sim b_k$
are required, then samples of future states and observations should be drawn from transition and observation models, respectively,
$x_{\tau}^{(i)} \sim \PT\left(\cdot \mid x_{\tau-1}^{(i)}, \pi_{\tau-1}(b_{\tau-1}^{(i)})\right)$,
$z_{\tau}^{(i)} \sim \PZ\left(\cdot \mid C^{(i)}, \xx_{\tau}^{(i)}\right)$.
Samples are denoted by bracketed index superscripts, $\square^{(i)}$.  
The estimation of the value function \eqref{eq: value function general reformulated} is given by
\begin{multline}
	\hat{V}_k^{(X,C)}\left(b_k, \pi_{k:L} \right) 
	= \\
	\frac{1}{N^s}
	\sum_{i=1}^{N^s}
	\sum_{t=k}^{L}
	r_t\left( a_t^{(i)}, C^{(i)}, \xx_t^{(i)}, b_t^{(i)}\left[C^{(i)}, \xx_t^{(i)}\right] \right)
	,
	\label{eq: value function approx X,C}
\end{multline}
where $b_t^{(i)}$ corresponds to history $H_t^{(i)}=(z_{1:t}^{(i)}, a_{0:t-1}^{(i)})$ and $a_t^{(i)} = \pi_t(b_t^{(i)})$.
Accordingly, the estimation of the objective function \eqref{eq: objective function state dep reward} assuming a state-dependent reward is given by
\begin{equation}
	\hat{J}_k^{(X,C)}\left(b_k, a_{k:L} \right)
	=
	\frac{1}{N^s}
	\sum_{i=1}^{N^s}
	\sum_{t=k}^{L}
	r_t\left( a_t, C^{(i)}, \xx_t^{(i)} \right)
	.
	\label{eq: objective function approx X,C}	
\end{equation}
A superscript $(X,C)$ in $\hat{V}_k^{(X,C)}$ is used to indicate estimators that use samples of both $\xx$ and $C$.
As will be shown in Section \ref{sec: samples from the original belief}, obtaining representative samples of the posterior belief at planning time is challenging.

For an explicit calculation of $C$, we can use the chain rule to formulate the value function as follows
\begin{align}
	&V_k\left(b_k, \pi_{k:L} \right) = \notag
	\\&=\!\!\!\!
	\ee{\belief{\xx_k}} \ee{\belief{C \mid \xx_k}}
	\left[
	\prod_{\tau=k+1}^L
	\ee{\PT\left(x_{\tau} \mid \cdots \right)}
	\ee{\PZ\left(z_{\tau} \mid \xx_{\tau}\right)}
	\!\left[
	\sum_{t=k}^{L}
	r_t\left( \cdot \right)
	\right]
	\right]
	\label{eq: value function X, C|X}
	\\&=\!\!\!\!	
	\ee{\belief{C}} \ee{\belief{\xx_k \mid C}}
	\left[
	\prod_{\tau=k+1}^L
	\ee{\PT\left(x_{\tau} \mid \cdots \right)}
	\ee{\PZ\left(z_{\tau} \mid \xx_{\tau}\right)}
	\!\left[
	\sum_{t=k}^{L}
	r_t\left( \cdot \right)
	\right]
	\right]
	\label{eq: value function C, X|C}
	.
\end{align}
Later we will use the corresponding estimation of \eqref{eq: value function X, C|X}, which is given by
\begin{align}\label{eq: expected reward explicit C}
		&\hat{V}_k^{(X)}\left(b_k, \pi_{k:L} \right) 
	=\\ 	
	&\sum_{i=1}^{N^s}\!
	\sum_{C \in \cc}  \!\frac{\belief{C | \xx_k^{(i)}}}{N^s}
	\sum_{t=k}^{L}
	r_t\!\left( a_t^{(i)}, C, \xx_t^{(i)}, b_t^{(i)}\!\left[C, \xx_t^{(i)}\!\right] \right)
	,\nonumber		
\end{align}
where $\xx_k^{(i)} \sim \belief{\xx_k}$. Since the expectation over $C$ is carried explicitly, samples $C^{(i)}$ are replaced by the variable $C$. All other variables are defined similarly to those in \eqref{eq: value function approx X,C}.
Consequently, the estimation of the objective function \eqref{eq: objective function state dep reward} is given by
\begin{equation}
	\hat{J}_k^{(X)}\left(b_k, a_{k:L} \right)
	=
	\sum_{i=1}^{N^s}
	\sum_{C \in \cc} \frac{\belief{C \mid \xx_k^{(i)}}}{N^s}
	\sum_{t=k}^{L}
	r_t\left( a_t^{(i)}, C, \xx_t^{(i)} \right)
	.
	\label{eq: objective function explicit C}
\end{equation}
In section \ref{sec:approach}, 
this formulation is further developed to calculate explicit expectations of $C$. 
Here we discuss the challenges in obtaining samples 
$\xx_k^{(i)}, C^{(i)} \sim b_k$ 
for the above estimators \eqref{eq: value function approx X,C},  \eqref{eq: objective function approx X,C}, \eqref{eq: expected reward explicit C},
and \eqref{eq: objective function explicit C}.

\subsubsection{Sampling Directly from the Belief} 
\label{sec: samples from the original belief}

Sampling from the hybrid belief $\belief{C, \xx_k}$ is challenging. Commonly, the belief is decomposed using chain rule into $\belief{C}\belief{\xx_k \mid C}$ \eqref{eq: value function C, X|C}, where $\belief{C}$ are the weights and $\belief{\xx_k \mid C}$ are the components.
Once the weights are calculated, we can sample hypothesis $C^{(i)} $ using the weights and sample $\xx_k^{(i)}$ using the component $\belief{\xx_k \mid C^{(i)}}$.
However, calculating all the weights is computationally prohibitive since the number of hypotheses is $|\cc| = (N^c)^{N^o}$, 
which is exponential complexity in the number of objects $N^o$. 
Additionally, it may be impossible to sample from the components $\belief{\xx_k \mid C}$ directly since not all probabilities can be sampled directly, and if the expectation of $\xx_k$ is intractable, the weights are also approximated 
$\belief{C} = \intop \belief{C, \xx_k}d\xx_k \approx \sum_i \belief{C, \xx_k^{(i)}}$, 
resulting in an additional layer of error.

Alternatively, the belief can be reformulated as $\belief{\xx_k}\belief{C \mid \xx_k}$ \eqref{eq: value function X, C|X}, allowing sampling $\xx_k^{(i)}$ from $\belief{\xx_k}$ and $C^{(i)}$ from $\belief{C \mid \xx_k^{(i)}}$.
Similarly to the previous case, samples of $\belief{\xx_k}$ cannot be obtained directly since it requires computing an exponential number of weights, 
\begin{equation}
	\belief{\xx_k} = \eta_{1:k} \sum_{C\in\cc} \belief{C} \belieftild{\xx_k \mid C}
	,
	\label{eq: normalized marginal X}
\end{equation}
and is not always possible to sample from the components $\belief{\xx_k \mid C}$.

However, there are MCMC methods to sample from a general multivariate distribution like MH \cite{Beichl00cse}. Yet, this requires querying the unnormalized marginal $\belieftild{\xx_k}$, given by
\begin{equation}
	\belieftild{\xx_k}
	=
	\sum_{C\in\cc}\belieftild{C, \xx_k}
	,
	\label{eq: unnormalized marginal X}
\end{equation}
which involves an \emph{exponential} number of components and is computationally prohibitively expensive. 
Querying belief means calculating the belief for a specific state value. Similarly, the unnormalized belief, the marginals, and the conditional beliefs can be queried.

Other MCMC methods like Gibbs sampling, require repeated sampling from $\belief{C \mid \xx_k}$ and $\belief{\xx_k \mid C}$, are viable but computationally expensive due to the large number of components involved. Despite its practicality, Gibbs sampling remains a costly alternative.

\subsubsection{Pruning Hypotheses}
Pruning hypotheses and keeping only a limited set is common practice, but it has severe drawbacks. Following pruning and renormalization, the belief is overconfident and the agent may assume that it knows the true hypothesis with very high probability while in fact the probability that the true hypothesis was pruned can be significant because most of the hypotheses were pruned.

Moreover, it is unclear how to keep the most probable hypotheses. To know which hypotheses are the most probable, it is necessary to calculate them. This requires calculating all hypotheses, which is intractable.

Additionally, the agent maintains a set of hypotheses at the beginning of the session. During operation, the agent received observations that may indicate that the true hypothesis was pruned. The agent cannot know whether the true hypothesis was pruned with a high probability since the probabilities of the pruned hypotheses are not calculated.

The un-normalized marginal belief $\belieftild{\xx_k}$ \eqref{eq: unnormalized marginal X}, can be approximated by marginalizing over a subset of hypotheses $\bar{\cc} \subset \cc$, which reduces complexity but introduces large estimation errors.
Our approach allows us to avoid such an estimation and query it explicitly.

\subsubsection{Importance Sampling}
\label{sec: importance sampling}
Important sampling (IS) is another common practice for estimating expectations. IS has two main problems in this setting. 
First, finding a suitable proposal distribution is challenging. 
Secondly, to calculate the importance ratio, we must compute the normalized belief and therefore the normalization factor $\eta_{1:k}$. 

Since, in general the expectation of $\xx_k$ is intractable, the normalization factor is also intractable. In this case the normalization factor can be approximate. One popular approximation is self-normalized importance sampling (SNIS), where samples obtained from the proposal distribution are used both for estimating the expectation and the normalization factor. The estimation mean square error (MSE) of SNIS is slightly larger than IS since the normalization factor is approximated \cite{Gabriel22nips}. 

Identifying an appropriate proposal distribution $q(\xx_k, C)$ that is easy to sample and provides accurate estimation is very challenging \cite{tokdar10sda}. The mean square error (MSE) of the IS and SNIS estimators are dependent on the proposal distribution, hybrid belief, and rewards. Generally the MSE calculation is intractable. 
Yet, Theorem \ref{theorem:exponential MSE of IS} will show that under certain assumptions, the MSE of IS can be bound from below by a bound that is exponential in the number of objects.

In comparison, if we could draw samples directly from the hybrid belief, the MSE would be exponentially smaller. Our approach empirically achieves accuracy similar to sampling from the original belief (see section \ref{Sec:experiment}).

Additionally, utilizing \TV{RBPF} \cite{Doucet00} requires marginalizing over $C$, which is computationally impractical.

\subsection{Explicit Expectation over $C$}

In an ambiguous environment, to ensure the agent operates safely and accurately, it is necessary to consider many states and possible hypotheses. In the case of sparse semantic probability $b_k[C]$, it is possible that the agent will be required to consider many hypotheses and will not be able to prune a significant number of hypotheses while maintaining the required level of accuracy and safety.

This motivates us to consider computing the value function \eqref{eq: expected reward explicit C} with an explicit expectation of $C$, i.e.~considering \emph{all} the hypotheses explicitly. There are, however, two principal difficulties associated with this challenge.
First, going through all the hypotheses is computationally prohibitive. 
Secondly, samples of $\xx_k$ are still required, which is computationally very costly if possible (see Section \ref{sec: samples from the original belief}).

%% file: approach.tex
\section{Approach \label{sec:approach}}
 
In section \ref{subsec: sampling belief}, we discussed the challenge of sampling representative samples of $\belief{\xx_k,C}$ and its significance for hybrid POMDP problems. 
Furthermore, the computational burden of querying $\belieftild{\xx_k}$ was discussed. 

In this section we show that $\belieftild{\xx_k}$ can be queried efficiently and \TV{that} representative samples can be drawn.
Next, we show that for a certain structure of state-dependent rewards, and \TV{under the assumption of} an open-loop setting, the expectation over $C$ can be calculated explicitly and efficiently, \TV{thereby} improving estimation accuracy.

\subsection{Planning-Time Belief Querying and Sampling}
\label{subsec:Samples from the belief}

In this section, the belief is reformulated to allow very efficient belief querying. Next, several sampling methods are discussed. These methods result in very accurate estimates of the objective function, as can be seen in Section \ref{Sec:experiment}.

In contrast to our method, sampling from the belief directly is intractable (see section \ref{sec: samples from the original belief}). Moreover, with from the shelf proposal distributions, importance sampling estimation may result in extremely high MSE, requiring an extensive number of samples \ref{sec: importance sampling}. 

The following definitions are required for Theorem \ref{lemma: belief formulation}.
Let the geometric belief $\geobelief{\xx_k}$ be the posterior probability of $\xx_k$ given previous actions $a_{1:k-1}$ and only geometric observations $z_{1:k}^g$,  
\begin{align}
	\geobelief{\xx_k} \triangleq \pr\left(\xx_k \mid z_{1:k}^g, a_{1:k-1}\right)
	.
	\label{eq:geometricbel}
\end{align}
The set of time indexes in which object $n$ has been observed until planning time \TV{$k$} is defined as $\Itime[n]$, for $n \in \left[1,\ldots,N^o\right]$.
The unnormalized conditional belief of $c_n \mid \xx_k$ is provided by 
\begin{equation}
	\belieftild{c_n \mid \xx_k}
= 
\pr_0\left(c_n\mid \xx_0\right)
\prod_{\tau \in \Itime[n] } 
\PZ\left(z_{\tau,n}^s \mid x_{\tau},x_n^o,c_n\right)
.
\label{eq: unnormalized belief c given X}
\end{equation}
\TV{In \eqref{eq: unnormalized belief c given X}, $z_{1:k}^g$ does not appear since it is independent of $c_n$ given $\xx_k$.}
Moreover, $\belieftild{c_n \mid \xx_k}$ can be conveniently normalized as follows, 
$
	\belief{c_n \mid \xx_k}
	\define
	\frac{
		\belieftild{c_n \mid \xx_k}
	}{
		\sum_{c_n=1}^{N^c} 
		\belieftild{c_n \mid \xx_k}
	}
	.
$

\begin{figure}
	\centering
	\includegraphics[width=0.35\textwidth]{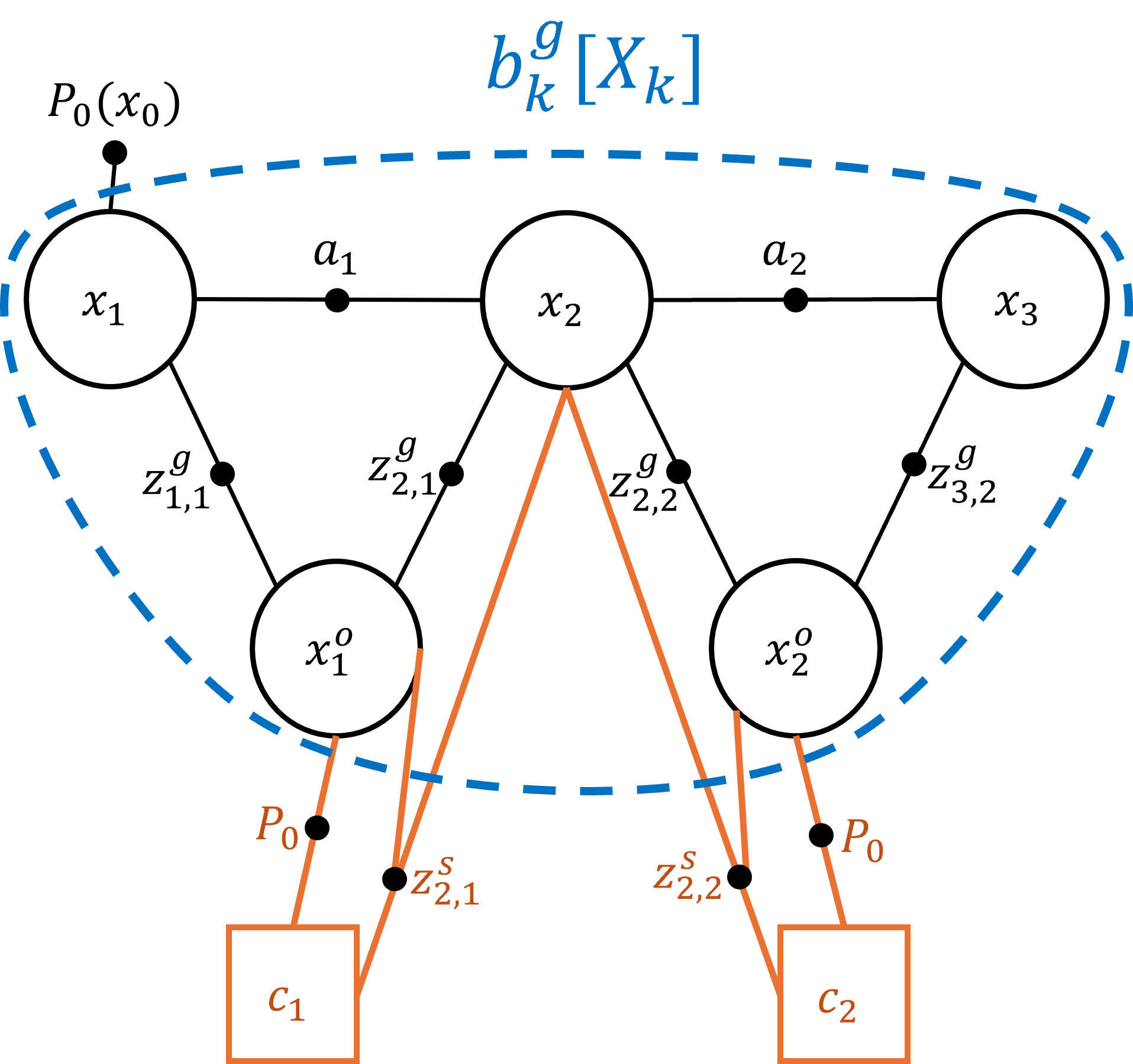}
	\caption{
	\TV{A factor graph diagram, representing the belief $b_k[C,\xx_k]$. The region outlined by the blue dotted line highlights the factor graph of the geometric belief $b_k^g[\xx_k]$, which corresponds to the geometric observations $z^g_{1:3}$, the actions $a_{1:2}$, and the robot prior probability $\pr_0(\xx_0)$. The semantic variables $c_1$ and $c_2$, in orange, are incorporated into the graph through semantic observations $z^s_{2, n}$ and semantic priors $\pr_0$. Given the state $\xx_k$, the semantic variables ($c_1, c_2$) are independent.}
	\label{fig:factor_graph_illustration}
	}
\end{figure}

\begin{theorem}
\label{lemma: belief formulation}
Let $\belief{\xx_k, C}$ be the hybrid belief \eqref{eq:belief} considering a viewpoint-dependent semantic observation model \eqref{eq: observation model}, and 
let $\geobelief{\xx_k}$ be the geometric belief \eqref{eq:geometricbel}.
Then, the hybrid belief $b_k$ can be formulated as follows
\begin{equation}
	\belief{C,\xx_k}
	=
	\eta_{1:k}
	\geobelief{\xx_k}
	\Phi\left(\xx_k\right)
	\prod_{n=1}^{N^o}
	\belief{c_n \mid \xx_k}
	\label{eq: belief X and C|X},
 \end{equation}
where $\eta_{1:k}$ is the normalization factor,
and 
\begin{equation}
	\Phi\left(\xx_k\right)
	\define
	\sum_{C\in\cc} 
	\prod_{n=1}^{N^o}
	\belieftild{c_n \mid \xx_k}
	\TV{=
	\prod_{n=1}^{N^o}
	\sum_{c_n=1}^{N^c}
	\belieftild{c_n \mid \xx_k}}
	.
	\label{eq: Phi}
\end{equation}
Furthermore, the \TV{unnormalized marginal belief 
$\belieftild{\xx_k}$ is given by}
\begin{equation}
	\TV{\belieftild{\xx_k}
	=
	\geobelief{\xx_k}
	\Phi\left(\xx_k\right) 
	\define \tilde{\eta}_{1:k} \belieftild{\xx_k}}.
	\label{eq: belief X}
\end{equation}
\end{theorem}
The proof of this theorem is provided in Appendix \ref{sec: proof of lemma belief formulation}.
\TV{The time complexity of querying $\belieftild{\xx_k}$ using formulation \eqref{eq: belief X} is $O(O_{b^g} \cdot N^c \cdot N^o)$, where $O_{b^g}$ is the time complexity of querying the geometric belief $\geobelief{\xx_k}$. This is because $\Phi\left(\xx_k\right)$ can be computed in $O(N^c \cdot N^o)$, according to \eqref{eq: Phi}. 
However, the time complexity of querying $\belieftild{\xx_k}$ in a brute force manner is $O\left(O_{b^g} \cdot (N^c)^{N^o}\right)$, since $\belieftild{\xx_k} = \sum_{C \in \cc} \tilde{b}_k [C,\xx_k]$ and $\cc$ has $(N^c)^{N^o}$ components. Querying $\belieftild{\xx_k}$ is required for sampling $X_k \sim \belief{\xx_k}$, as will be further discussed in this section. Equation \eqref{eq: belief X} enables efficient querying of $\belieftild{\xx_k}$, through exploitation of the conditional independence of the semantic variables given the state $\xx_k$.}
\TV{Figure \ref{fig:factor_graph_illustration} illustrates the factor graph representing the  belief $b_k[C,\xx_k]$ and outlining the conditional independence of the semantic variables.}

We now analyze the impact of Theorem \ref{lemma: belief formulation} considering two common methods for generating samples and estimating the expectation of some function $g(X_k,C)$: MCMC sampling and IS estimation. In our context, the function $g(\xx_k,C)$ is the integrand of the value function \eqref{eq: value function general reformulated} or integrand of the  objective function \eqref{eq: objective function state dep reward}.


Both approaches require querying the unnormalized marginal belief $\belieftild{\xx_k}$ \eqref{eq: unnormalized marginal X}, i.e. to evaluate it for a given realization of $\xx_k$. This evaluation is computationally expensive
 without Theorem \ref{lemma: belief formulation}.

 However, using  \eqref{eq: belief X}, it is possible to query the unnormalized marginal belief $\belieftild{\xx_k}$ efficiently.  The semantic component $\Phi\left(\xx_k\right)$  is the key to efficient computation. According to \eqref{eq: belief X}, $\Phi$ consists of $(N^c)^{N^o}$ components, but can be queried in a running time of $O(N^c \cdot N^o)$. It is the reason we can query the belief efficiently.

To draw samples from $\belief{\xx_k}$, MCMC sampling methods such as \cite{Ihler03nips} and MH \cite{Beichl00cse} can therefore be used since efficient querying of $\belieftild{\xx_k}$ is possible. 
Because MCMC samples are drawn in an iterative manner, the algorithm adds additional computational complexity. We will consider $N^{it}$ iterations of the MCMC algorithm, which will be included to the complexity of the algorithm.
Contrary to this, drawing samples directly from $\belief{\xx_k}$ is intractable since it has exponentially many components, as discussed in section \ref{sec: samples from the original belief}.

Alternatively, we can use IS estimation. However, finding a suitable proposal distribution $q(C,X_k)$ can be challenging. 
Next in Theorem \ref{theorem:exponential MSE of IS} we show that under Assumptions \ref{assumption:nondegen}-\ref{assumption:marginal}, the IS estimation MSE  is exponential in the number of objects $N^o$. 
Then, we will propose a proposal distribution of the form $q(\xx_k, C) = q(\xx_k)\belieftild{C\mid \xx_k}$ and show that the MSE of the corresponding IS estimator is not bound by an exponential bound and can achieve much smaller MSE. As earlier, Theorem \ref{lemma: belief formulation} will be the key to performing calculations efficiently.
For Assumption \ref{assumption:consistent}, define $C^{tr}$ to be the true hypothesis. $C^{tr}$ is a realization of the prior probability, thus $C^{tr} \sim \pr_0(C)$.

\begin{assumption}
	\label {assumption:nondegen}
	The marginal prior $\pr_0(C)$, is non-degenerate, i.e. $\pr_0(C) > 0$ for all $C \in \cc$. 
\end{assumption}
\begin{assumption}
	\label {assumption:consistent}
	\TV{$C \mid H_k$ is weakly converging to $C^{tr}$ in probability. Thus,
	for all $C^{tr} \in \cc$ and for all $\epsilon > 0$, there exists $k_0$ such that for all $k > k_0$,
	$
	\pr \left((C \mid H_k) = C^{tr} \right) \geq 1-\epsilon
	$.}
\end{assumption}
\begin{assumption}
	\label {assumption:marginal}
	For any proposal distribution $q(C,\xx_k)$, its marginal $q(C)$ is independent on $H_k$ and $C^{tr}$.
\end{assumption}

\TV{Assumption \ref{assumption:consistent} means that as the robot gathers more observations, the probability to sample true hypothesis $C^{tr}$ goes to 1.} Furthermore, this assumption can be relaxed to assume that the belief is asymptotically \TV{concentrated} around a small number of hypotheses, $\cc_{sub} \subset \cc$, such that \TV{$\lim_{k \to \infty} \pr \left(C \in \cc_{sub}\right) = 1$}.
Each realization of $C^{tr}$, can lead to a different observations sequence, and therefore different history space 
$\mathcal{H}_k \left(C^{tr}\right)$.

\begin{theorem}
	\label{theorem:exponential MSE of IS}
Let $\pr_0\left(C,\xx\right)$, $q(C,\xx_k)$, and $\belief{\xx_k,C}$ satisfy Assumptions \ref{assumption:nondegen}-\ref{assumption:marginal}. Consider estimating the expectation of some function $g\left(\xx_k,C\right)$ using an IS estimator denoted by $\hat{\E}_{IS}^{\left(\xx,C\right)}$. 
Define 
$\bar{g}\left(C\right)$
to be the conditional expectation of $g$, 
$
\bar{g}\left(C\right)
\define
\ee{\xx_k\mid C} 
\left[g\left(\xx_k,C\right)\mid C\right]
$,
and 
$
w(C)
\define 
\eta_w\frac{\pr_0(C)}{q(C)}
$
a new distribution with the normalizer $\eta_w$. Then \TV{as $k$ goes to infinity,} the MSE of
$\hat{\E}_{IS}^{\left(\xx,C\right)}$ can be bounded from below by 
\begin{equation}
\mse\left(\hat{\E}_{IS}^{\left(\xx,C\right)}\right)
\geq
\frac{1}{N^s}
\left|\cc\right|\underset{w}{\var}\left(\bar{g}\right)
\label{eq:exponential MSE}
.
\end{equation}
\end{theorem}
The proof is provided in appendix \ref{sec: appendix important sampling illustration}.

Since $\left|\cc\right|$ is exponential in $N^o$,
the MSE (\ref{eq:exponential MSE}) is exponential.
\TV{
For example, if the proposed distribution of $C$ is uniform, as $k$ increases and more observations are collected, the semantic belief $b_k[C]$ is expected to be increasingly concentrated around specific hypotheses. Consequently the MSE will increase with the time-step $k$.
}
As a consequence of \TV{Theorem} \ref{theorem:exponential MSE of IS}, any proposal distribution should be evaluated in terms of estimation MSE, since scenarios with proposal distribution that fulfill Assumptions \ref{assumption:nondegen}-\ref{assumption:marginal}, will result in exponential MSE. 
Estimators with exponential MSEs will be highly inaccurate and unreliable. In such a scenario, reducing the MSE to a reasonable level will require an exponential number of samples.

In contrast, by using a proposal distribution of the form 
$q(\xx_k, C) = q(\xx_k)\belieftild{C\mid \xx_k}$ estimation accuracy can be improved.
According to \cite{tokdar10sda}, the MSE of SNIS is  proportional to the variance of the importance weights.
The importance weights for the above proposal distribution are given by 
\begin{equation}
	\omega_{\text{IS}}(\xx_k, C)
	= 
	\frac{\belieftild{\xx_k, C}}{q(\xx_k, C)} 
	=
	\frac{\belieftild{\xx_k}}{q(\xx_k)}
	.
	\label{eq: IS weights}
\end{equation}
Since $\belieftild{C \mid \xx_k}$ does not affect importance weights, they do not increase the MSE. 
Furthermore, following Theorem \ref{theorem:exponential MSE of IS} and bounding the MSE from below using inequality \eqref{eq:IS of C smaller then IS of C and X}, then the bound on the MSE using our proposal distribution 
$q(\xx_k, C) = q(\xx_k)\belieftild{C\mid \xx_k}$
will be 
\begin{equation}
	\mse\left(\hat{\E}_{IS}^{\left(\xx,C\right)}\right)
	\geq
	\frac{1}{N^s}
	\underset{\belief{C}}{\var}\left[\bar{g}\right]
	.
	\label{eq:bound of our SNIS}
\end{equation}
In this case the MSE is not exponential and therefore exponentially smaller than in Theorem \ref{theorem:exponential MSE of IS}. 

Computing $\omega_{\text{IS}}$ for our proposal distribution involves querying $\belieftild{\xx_k}$, which is only possible using Theorem \ref{lemma: belief formulation}, otherwise, the computational cost \TV{is exponential}.
A reasonable proposal distribution candidate for $q(\xx_k)$ would be $\geobelief{\xx_k}$, since it already incorporates geometric history. This resulted in $\omega_{\text{IS}} = \Phi\left(\xx_k\right)$.



\subsection{Structured State Dependent Reward}
\label{sec: structured reward}

To increase estimation accuracy, instead of taking samples of $C$, we can compute the expectation over $C$ explicitly. The Rao-Blackwell theorem \cite{Blackwell47ams} states that this will improve the estimation accuracy. Furthermore, because all semantic mappings are taken into account in the expectation, the true semantic mapping is considered.

\TV{
Methods developed in this section are applicable to open-loop planning and policy evaluation, i.e.~for any given policy. 
Extending these results to closed-loop planning is left for future work.
}

\begin{theorem}[Rao-Blackwellization for Sampled Based Estimations]
	\label{theorem: Rao-Blackwell for samples}  
	Let $\xx$ and $C$ be random vectors, with the joint probability density function $b\left[X,C\right]$, domain $\Xspace \times \cc$, marginal density $b[\xx]$ and conditional $b\left[C\mid \xx\right]$.
	Let $\left\{\xx^{(i)}, C^{(i)}\right\}_{i=1}^{N^s}$ be a set of $N^s$ independent and identically distributed (iid) samples drawn from $b\left[X,C\right]$.
	Let $g:\Xspace \times \cc \to \mathbb{R} $ be a function from $\Xspace \times \cc$ to the real numbers $\mathbb{R}$.
	Define an estimator of $\E\left[g(\xx,C)\right]$ using samples 
	$\left\{\xx^{(i)}, C^{(i)}\right\}_{i=1}^{N^s}$
	as follows
	\begin{equation}
		\hat{E}^{(\xx,C)} \left[g\right] = \frac{1}{N^s} \sum_{i=1}^{N^s} g(\xx^{(i)},C^{(i)})
		.
	\end{equation}
	Define another estimator using only samples 
	$\left\{\xx^{(i)}\right\}_{i=1}^{N^s}$
	and explicitly calculated expectation over $C \mid \xx^{(i)}$ by
	\begin{equation}
		\hat{E}^{(\xx)} \left[g\right] 
		=
		\frac{1}{N^s} \sum_{i=1}^{N^s} 
		\sum_{C\in\cc} g(\xx^{(i)},C) b\left[C \mid \xx^{(i)}\right]
		.
		\label{eq:estimator explicit C}
	\end{equation}
Then  
$\mse(\hat{E}^{(\xx)}) \leq \mse(\hat{E}^{(\xx, C)})$,
with the relation between the MSEs given by
\begin{multline}
\mse(\hat{E}^{(\xx)}) = \mse(\hat{E}^{(\xx, C)})
-
\frac{
\ee{b[\xx]} 
\left[
	\underset{b\left[C\mid \xx \right]}{\var}
	\left\{
		g \left( \xx,C \right) \mid \xx 
	\right\}
\right]}
{N^s}
. 		
\label{eq: Rao-Blackwell explicit C}
\end{multline}
Since the variance
$
\underset{b\left[C\mid \xx \right]}{\var}
\left(
	g \left( \xx,C \right) \mid \xx 
\right)
$
is non-negative, equality holds if and only if the variance is zero for all non-negligible subsets of $\Xspace$.
\end{theorem}


The proof is provided in Appendix \ref{proof: Rao-Blackwell for samples}.
This is a special case of the Rao-Blackwell theorem for sample based estimators \TV{which was  added here for the completeness of the mathematical presentation. In this special case, it is possible to calculate a theoretical term for the MSE reduction as shown in \eqref{eq: Rao-Blackwell explicit C}.
In certain special cases that will be discussed later, the Rao-Blackwell theorem can be applied to improve the estimation accuracy of the value function \eqref{eq: value function approx X,C} or the objective function \eqref{eq: objective function approx X,C}. Since all hypotheses are considered in such cases, the true hypothesis is always taken into account. However, when pruning or sampling hypotheses this cannot be guaranteed.
} 

\begin{lemma}  
\label{lemma: expected reward brute force}
Let $\xx_t, H_k$, and $\pi_{k:t-1}$ denote the state at time-step $t$, the history at time-step $k$, and future \TV{policy} sequence, respectively. 
\TV{Assume that the inner reward function is state-dependent and it is not directly dependent on the policy $r_t = r_t(\xx_t, C)$, or, it is dependent on the action in an open-loop setting.  
} Then the value function can be formulated as follows
\begin{equation}
	V_k\left(b_k, \pi_{k:L} \right) 
	=
	\ee{\pr\left(\xx_L \mid H_k, \pi_{k:L-1}\right)}
	\left[
	\ee{\belief{C \mid \xx_k}}
	\left[
	\sum_{t=k}^{L}
	r_{t}\left(\cdot\right)
	\right]
	\right]
	\label{eq: value function brute force}
	.
\end{equation}
\label{lemma: expected reward state dependent}
\end{lemma}
The proof of Lemma \ref{lemma: expected reward brute force} is provided in Appendix \ref{proof: expected reward brute force}.

\TV{Following Lemma \ref{lemma: expected reward brute force}, the expected reward of one reward element $\rho_t$ is given by}
\begin{multline}
	\ee{z_{k+1:t}}\left[\rho_t\left(b_t\right)\right]
	=
	\ee{\pr\left(\xx_t \mid H_k, \pi_{k:t-1}\right)}
	\left[
		\ee{\belief{C \mid \xx_k}}
		\left[
			r_{t}\left(\cdot\right)
		\right]
	\right]
	.
	\label{eq: expected reward brute force}
\end{multline}
Accordingly, the estimation of \eqref{eq: expected reward brute force} using samples $\left\{\xx_t^{(i)}\right\}_{i=1}^{N^s} \sim \pr\left(\xx_t \mid H_k, \pi_{k:t-1}\right)$ and explicit expectation over $C$ is given by
\begin{equation}
	\hat{\E}^{(\xx)}[\rho_t(b_t)]
	=
	\frac{1}{N^s}
	\sum_{i=1}^{N^s}
	\sum_{C \in \cc} \belief{C \mid \xx_k^{(i)}}
	r_t\left(\cdot\right)
	.
	\label{eq: expected reward approx brute force}
\end{equation}
Since the number of semantic hypotheses is $(N^c)^{N^o}$ the computation complexity of \eqref{eq: expected reward approx brute force}
is $O\left(N^s \cdot (N^c)^{N^o} O_r\right)$, where $O_r$ is the computational complexity of calculating $r$. 

\TV{
The computation of Equations~\eqref{eq: value function brute force}--\eqref{eq: expected reward approx brute force} is possible only in open-loop or policy evaluation settings, since the expectation applied assumes that the policy sequence $\pi_{k:t-1}$ is given. Applying explicit expectation over $C$ in a closed-loop setting for policy optimization necessitates generating future observations for all hypotheses, which is computationally infeasible. 
Nevertheless, sampling observations without hypothesis pruning (Section~\ref{subsec:Samples from the belief}) remains feasible in  a closed-loop setting.
}

As another major contribution of this study, we show that it is possible to calculate the objective function \eqref{eq: objective function state dep reward}  with an explicit expectation over $C$ very efficiently, for a special structured reward function. This improves estimation error.

\subsubsection{Efficient Computation for Structured State Dependent Rewards}

Define $\theta \define \{1,\ldots, N^o\}$ as the set of all objects' indexes, and define the following structure of inner reward functions
\begin{equation}
r_t \left(C,\xx_t \right)
=
\sum_{j=1}^{|\Theta|} \prod_{n \in \theta_j}
	r_{t,j,n}\left(c_n, \xx_t\right)
,
\label{eq: efficient reward}
\end{equation}
where ${\theta_j}$ is a subset of the objects' indexes, 
$\theta_j \subseteq \theta $, and $\Theta$ is a set of such subsets. 
$\Theta$ is a subset of the power set of $\theta$, $\Theta \subseteq \mathcal{P}(\theta)$. It is assumed that $|\Theta|$ is small, $|\Theta| \ll |\mathcal{P}(\theta)|$.
For simplicity, the number of inner reward elements $r_{t,j,n}$ participating in the inner reward function $r_t$ \eqref{eq: efficient reward} denoted by $\Thetasize = \sum_{j=1}^{|\Theta|}|\theta_j|$.
The computation complexity of a structured inner reward function is  $O_r = O(\Thetasize)$.

Following are special cases of the structured state-dependent reward \eqref{eq: efficient reward}:
\begin{align}
r_t\left(C,\xx_t\right) & 
	=\sum_{n \in \theta} r_{t,n}\left(c_n, \xx_t\right),
	\label{eq:state reward add}\\
r_t\left(C,\xx_t\right) & 
	=\prod_{n \in \theta} r_{t,n} \left(c_n, \xx_t\right),
	\label{eq:state reward prod}
\end{align}
where \eqref{eq:state reward add} is referred as an additive reward and \eqref{eq:state reward prod} 
as a multiplicative reward.
The additive reward is a special case where
$\Theta = \{\theta_j\}_{j\in\theta}, \;\;\theta_j = \{j\}$,
and the multiplicative reward \eqref{eq:state reward prod} 
is a special case where $\Theta$ has only one element: $\theta$.

An example of an additive reward is object search.
Object search refers to a problem in which the robot attempts to find any object of a particular class type. 
This can be formulated using the additive reward, where the distance to a specific class type is minimized.
It will be shown that $\psafe$ is a special case of multiplicative rewards.


\begin{theorem}
\label{theorem: expected reward efficient}
Let the reward be of a structured reward function \eqref{eq: efficient reward}.
The expected reward \eqref{eq: expected reward brute force} can be reformulated as follows
\begin{multline}
	\ee{z_{k+1:t}}\left[\rho_t\left(b_t\right)\right]
	=\\
	\ee{\pr \left( \xx_\TV{t} \mid H_k, \pi_{k:t-1} \right)}
	\left[
		\sum_{j=1}^{|\Theta|}
		\prod_{n \in \theta_j}
		\ee{\belief{c_n \mid \xx_k}}
		\left[
		r_{t,j,n}\left(c_n,\xx_t\right)
		\right]
	\right]
	,
	\label{eq:expected value state dependent final}
\end{multline}
and the corresponding estimator is given by
\begin{multline}
	\hat{\E}^{(\xx)}[\rho_t(b_t)]
	=\\
	\frac{1}{N^s}
	\sum_{i=1}^{N^s}
	\sum_{j=1}^{|\Theta|}
	\prod_{n\in \theta_j}
	\sum_{c_n=1}^{N^c}
	\belief{c_n \mid \xx_k^{(i)}}
	r_{t,j,n}\left(c_n,\xx_t^{(i)}\right)
	.
	\label{eq:expected reward efficient approx}
\end{multline}
Computational complexity of \eqref{eq:expected reward efficient approx} is  
$O\left(N^s \Thetasize N^c \right)$.
\end{theorem}


The proof of Theorem \ref{theorem: expected reward efficient} is provided in the Appendix \ref{sec: proof of theorem expected reward efficient}.
Equations \eqref{eq:expected reward efficient approx} and \eqref{eq: expected reward approx brute force} provide the same exact estimation. Thus, if the same samples are used, the results will be the same. 
Yet, the computational cost of 
\eqref{eq:expected reward efficient approx}
is $O\left(N^s \Thetasize N^c \right)$, 
while the naive brute force method 
\eqref{eq: expected reward approx brute force}
complexity is 
$O\left(N^s \Thetasize \cdot (N^c)^{N^o}\right)$,
resulting in a significant reduction in the computational cost.

A combined SNIS estimation with an explicit expectation over $C$ can be achieved by using samples of $\xx_k^{(i)} \sim q(\xx_k)$ to calculate the approximated expected reward \eqref{eq:expected reward efficient approx}.
In this case both the weight \eqref{eq: IS weights} and the explicit expectation over $C$ can be calculated efficiently using our algorithm.

\subsubsection{Numerical Example}
\label{sub: structured reward / numerical example}

Consider the following scenario. 
$N^c = 1000$ types of classes, 
$N^o = 10$ number of detected objects, 
and $N^s = 100$ number of samples. 
The robot's objective is to search for objects of a specific type, let's assume a baseball.
The inner reward function can be formulated as
\begin{equation}
	r_{t,n}(c_n, \xx_t)
	=
	\oneb \left[c_n = \text{baseball}\right]
	|| x_o^n - x_t ||^2,
\end{equation}
\TV{where $\oneb \left[\cdot \right]$ is the indicator function.}
Then $ \Thetasize = N^o$.
In a brute force approach \eqref{eq: expected reward approx brute force},  computational complexity is
\begin{align}
	O(\hat{\E}^{(\xx)}[\rho_t(b_t)]) &= O\left(N^s \cdot (N^c)^{N^o} \Thetasize\right), 
\end{align}
which is impractical for real-time applications.

The same samples $\left\{\xx_t^{(i)}\right\}_{i=1}^{N^s}$, will yield the same numerical result using our method \eqref{eq:expected reward efficient approx}.
However, the computational complexity is
\begin{align}
	O(\hat{\E}^{(\xx)}[\rho_t(b_t)]) &= O\left(N^s \Thetasize N^c \right),
\end{align}
which is a significant reduction in running time.

A question that may arise is why the number of objects $N^o$, does not affect computational complexity. It can hide inside the set $\Theta$, affecting the computational complexity. Moreover, if an object does not participate in the reward function, it is automatically marginalized in our method.

\subsection{Structured-Reward Representation of $\psafe$}
\label{sub: structured reward psafe}

In general, the evaluation of $\psafe$ \eqref{eq:psafe} must be as accurate as possible. According to the Rao-Blackwell theorem, we can estimate it with explicit expectations of $C$, thereby increasing its accuracy. Here we will show that $\psafe$ has the same structure as the value function with multiplicative reward \eqref{eq:state reward prod}. Assuming an open loop setting, we can estimate $\psafe$ very efficiently with an explicit expectation  over $C$, similarly to \eqref{eq:expected reward efficient approx}.

\begin{proposition}
	\label{prop: safety inner reward}
	
	$\psafe$ from \eqref{eq:psafe} can be expressed as the expected reward with the inner reward function
	\begin{equation}
	r_{safe}\left(\xx_L, C\right) 
	=
	\prod_{n=1}^{N^o}
	\prod_{t=k+1}^L
	\oneb \left[x_{t} \notin \xunsafe \left(c_n, x^o_n\right) \right]
	\label{eq: safety inner reward}
	.
	\end{equation}
	Consequently, $\psafe$ is given by
	\begin{equation}
	\psafe
	=
	\ee{z_{k:L}} 
	\left[
	\ee{b_L \left[\xx_L,C\right] }
	\left[
		r_{safe}\left(\xx_t, C\right)
	\right]
	\mid b_k, \pi_{k:L} 
	\right]
	.
	\label{eq: psafe as expected reward}
	\end{equation}
\end{proposition}	

The proof of proposition \ref{prop: safety inner reward} is provided in the appendix \ref{proof: safety inner reward}.
\eqref{eq: psafe as expected reward} is a special case of the multiplicative reward \eqref{eq:state reward prod}, with element of the reward given by
\begin{equation}
	r_{L,n}\left(c_n, \xx_t\right)
	=
	\prod_{\tau = k+1}^{L}
		\oneb\left[x_{\tau} \notin \xunsafe\left(c_n,x^o_n\right)\right]
	\label{eq:safety reward element}.
\end{equation}
In this case, $\Thetasize = N^o$, and the computational runtime is $O\left(N^s N^o N^c\right)$.

Using our method to calculate the probability of safety, we account for all possible semantic mappings, since our method is equivalent to explicitly considering all possible semantic mappings, achieving better estimation than pruning or sampling hypotheses.

\input{table_methods.tex}

%% file: table_methods.tex
\begin{table*}
\centering
\begin{tabular}{
|p{3.1cm}||p{0.9cm}|p{1.5cm}|p{5.2cm}|p{1.5cm}|  }
\hline
Belief representation&  	Accurate&      Probability guarantees (Hoeffding)&     Incremental runtime & non-linear general models\\
\hline
\hline
\Ebfull&	        V&	V&	Exponential - $O\left((N^c)^{N^o} \cdot N^s \cdot N^\theta\right)$&		                X\\
\Ebpruned&	        X&	X&	Constant -  $O\left(|\cc^{sub}| \cdot N^s \cdot N^\theta\right)$&		                X\\
\PFfull&	        V&	X&	Exponential - $O\left((N^c)^{N^o} \cdot N^s \cdot N^\theta\right)$&		                V\\
\PFpruned&          X&	X&	Constant -  $O\left(|\cc^{sub}| \cdot N^s \cdot N^\theta\right)$&    	                V\\
\textbf{\MCMC}&		V&	V&	Polynomial- $O\left(k\cdot N^c \cdot N^o \cdot N^s \cdot N^\theta \cdot N^{it}\right)$&	V\\
\textbf{\SNIS}&		V&	X&	Polynomial - $O\left(k\cdot N^c \cdot N^o \cdot N^s \cdot N^\theta \right)$&		    V\\
\hline
\end{tabular}
\caption{
Comparison of different methods. The runtime refers to the time required to update the belief/samples in response to taking one action and receiving one observation.
\texttt{MCMC(Ours)}  estimator is both accurate and reasonably efficient for real-time applications. It can provide probability guarantees such as Hoffding's inequality.
In contrast, \texttt{SNIS (ours)} is less accurate but more rapid. Asymptotically, it is unbiased, and Hoffman's inequality does not apply to it. 
\texttt{Exact-all-hyp} complexity is exponential. Since the samples are drawn from the theoretical belief, it is the most accurate estimate in this study. Generally, the computation of the theoretical belief and drawing samples from it is intractable.
Utilizing \texttt{PF-all-hyp} results in an exponential runtime and therefore is unpractical. This estimator does not provide probabilistic guarantees such as Hoffding inequality. 
\texttt{PF-pruned} and \texttt{Exact-pruned} solve the complexity of their full hypotheses versions. However, they result in large errors.
}
\label{table: methods performance}
\end{table*}

%% file: experiments.tex
\section{Experiments \label{Sec:experiment}}


Our methods were evaluated using a Python simulation \TV{in} a synthetic 2D environment. \TV{The} primary \TV{claim} is that our method can estimate the objective function, \TV{for instance,} the probability of safety accurately and efficiently. In contrast, other methods cannot \TV{achieve} both. To verify \TV{this} claim, we designed an experiment in which the belief can be calculated analytically without approximations or estimations.  Despite this, the expected reward and the probability of safety do not have analytical solutions and \TV{are} therefore approximated \TV{using} samples.

To obtain an analytical solution for belief propagation, we assume that all prior probabilities of \TV{the} continuous variables are  Gaussian \TV{mixture model (GMM)} and that the observation and transition models are linear and Gaussian . For each class type, a different observation model \TV{is assumed}. The result is a hybrid belief that contains semantic discrete variables $C$ \TV{and} continuous state variables $\xx_k$\TV{, all of which} are interconnected. In this case\TV{, the analytical solution for the propagated} belief of $\xx_k$ is a GMM, \TV{and} the GMM weights are the marginal belief of $C$. An analytical solution to belief propagation \TV{can be found in \cite{Alspach72tac}. Due to computational limitations, this solution can be applied only in scenarios with} small numbers of objects and classes, since the number of hypotheses is combinatorial.

We considered an open-loop setting \TV{in which only} predefined action sequences \TV{are considered}. This setting \TV{enables evaluation of} objective function estimators in a controlled scenario.

\subsection{Simulation Setting}

Our simulated scenario consists of a robot traveling in a 2D environment with scattered objects. The $n$th object is represented by a location $x_n^o$ and a class $c_n$. In each simulation, the class and location of each object are chosen randomly according to their respective prior probabilities. Each object also has an unsafe area, defined by its class and centered at the object's location.

Starting at the origin, the robot either moves according to a predefined sequence of actions (Section \ref{sec:Res_GivenActionSeq}), 
or it performs planning and chooses the best action sequence from a given set of candidate action sequences (Section \ref{sec:Res_Planning}).  
  At each time-step, the robot receives a geometric and a semantic observation from each object. 
The geometric observation model is given by 
$
z_{k,n}^g = x_n^o - x_k + v_{k,n}^g
$, 
where $ v_{k,n}^g \sim \mathcal{N}(0, \eye\sigma^2 )$ and $ \sigma^2 = 5$. The semantic observation model is given by 
$
z_{k,n}^s 
=
\alpha_{c_n}\left(x_n^o - x_k\right) + v_{k,n}^s
$, 
where $ v_{k,n}^s \sim \mathcal{N}(0, \eye\sigma^2 )$ 
and 
$ \alpha_{c_n}$ is equally spaced between 
$\alpha_{c_n} = 0.95$ for $ c_n = 1 $ 
and 
$\alpha_{c_n} = 1.05$ for $ c_n = N^c $.
The transition model is given by $ x_{k+1} = x_k + a_k + w_k$, where 
$ w_k \sim \mathcal{N}(0, \eye\sigma^2_x )$, 
$ \sigma^2_x = 0.3 $. 
The observation noises and the process noise $v_{k,n}^g, v_{k,n}^s$ and $w_k$, are independent on each other and on noises of different time-steps.
  
Since the robot has only partial knowledge of the environment represented by the prior probability $\pr_0(\xx_k, C)$, it infers the environment using the observations it receives. Following inference, the robot estimates the probability of safety $\psafe$ and the expected reward.




We compare our estimation methods to the following estimation methods. Samples of $\xx_k$  are drawn from the theoretical belief, followed by an explicit expectation over the semantic hypotheses $C$. This estimation method will be referred to as the \ebfull. This method is the most accurate and we do not claim our method is more accurate. However, this method requires an analytical solution to belief propagation, which is not available in the general case, and it runs explicitly over all hypotheses which is computationally very expensive. 

Another method is using a particle filter. For each hypothesis we can consider representing conditional beliefs $\belief{\xx_k \mid C}$, by a particle filter followed by estimation of $\belief{C}$ using \eqref{eq:belief weight recursive} and an explicit expectation over hypotheses $C$. This method does not require an analytical solution for belief propagation. However, the computational complexity of running over all hypotheses remains. Since all hypotheses are explicitly considered, we will refer to this method as \PFfull.

To deal with the exponential number of hypotheses, we will modify these two methods by pruning all hypotheses except three. These methods are referred to as \ebpruned and \PFpruned, respectively. These methods will be given an advantage in our experiments by keeping the hypotheses with the highest probability, which is typically not provided.

For our methods, we will consider the MCMC method, where samples from the belief are approximated using the MH algorithm, and SNIS method, in which samples from the geometric belief are drawn from $b_k^g(\xx_k)$ and weighted according to \eqref{eq: IS weights}.
These methods will be referred to as \MCMC and \SNIS, respectively.

Lastly, the geometric and semantic components are considered separately by taking the MAP estimator , $\xx_k^{MAP} = \argmax{\xx_k}{\geobelief{\xx_k}}$, followed by the MAP estimator of $C\mid X$, $C^{MAP} = \argmax{C}{\belief{C \mid \xx_k^{MAP}}}$. We will refer to this method as \GSMAP - geometric semantic MAP.

\subsection{Simulation Results for a Pre-defined Action Sequence}\label{sec:Res_GivenActionSeq}

\begin{figure}
	\begin{subfigure}[b]{\linewidth}
	\centering
	\includegraphics[width=0.75\linewidth]{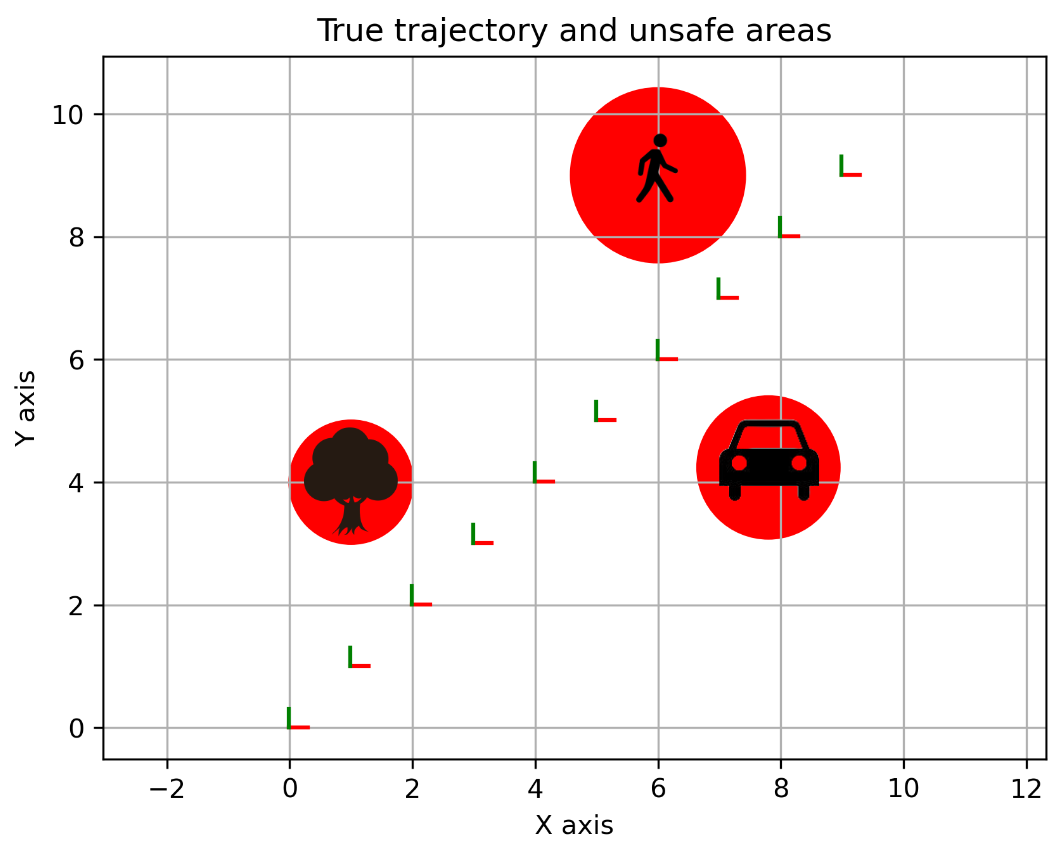}
	\caption{}
	\label{fig: sim1 scene properties}
	\end{subfigure}
	
	\begin{subfigure}[b]{\linewidth}	
	\centering
	\includegraphics[width=0.70\linewidth]{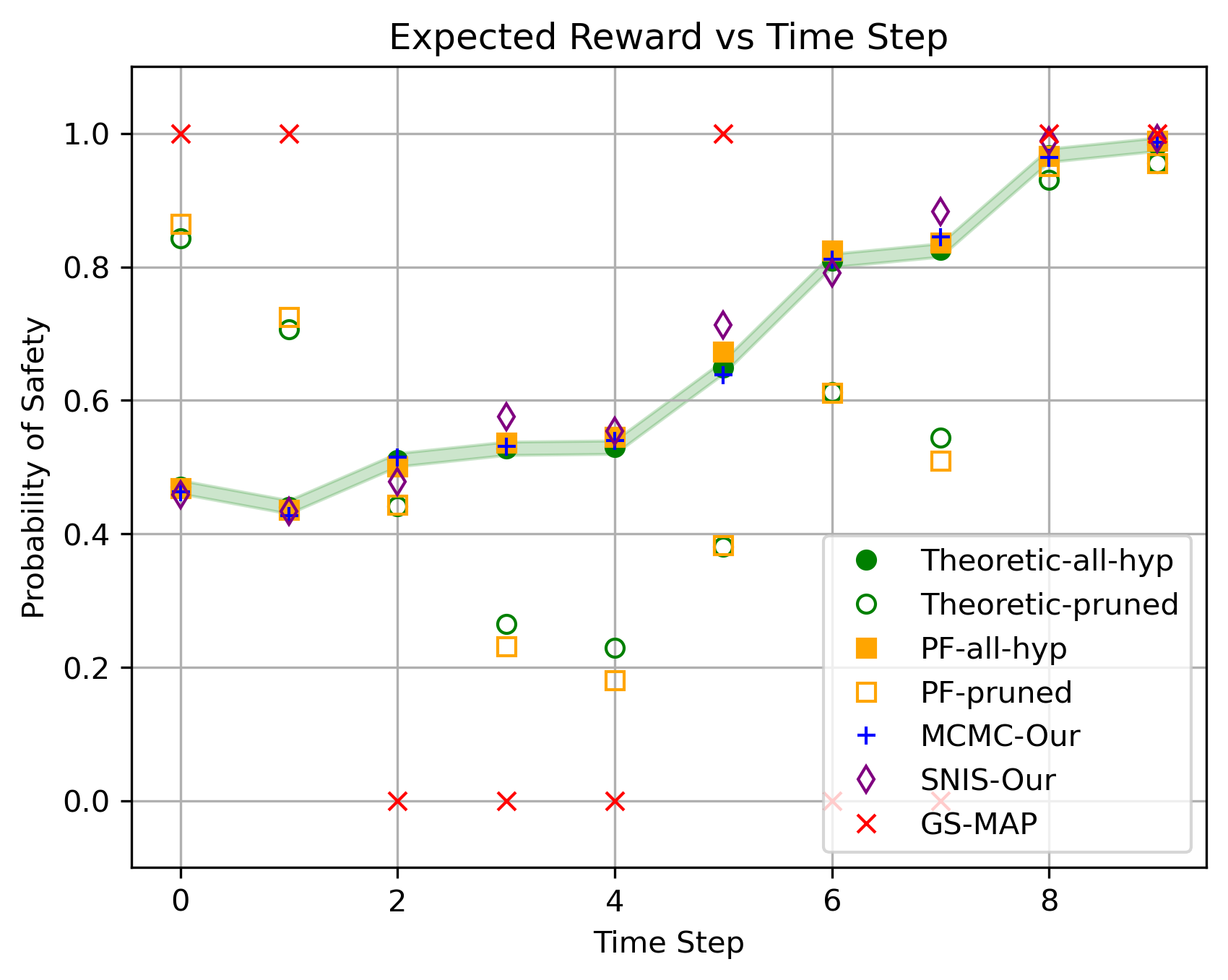}
	\caption{}
	\label{fig: sim1 expected reward}
	\end{subfigure}
	\caption{\textbf{(a)} The true trajectory and objects' unsafe area. Robot starts at (0,0) and reaches (9,9). The class of an object specifies the unsafe area around it. Each time step, $\psafe$ is calculated assuming the robot will continue to move in a straight line. \textbf{(b)} One trial probability of safety versus time, compared between different methods. In the case of the \ebfull, Hoffding's inequality holds, ensuring this result. For the rest of the estimators, \MCMC and \PFfull are aligned with the \ebfull, \SNIS is slightly further away, and \ebpruned, \PFpruned and \GSMAP are more biased. The bias can be positive or negative, and there is no indication as to which it will be. In the case of \GSMAP we get the least accurate results.}
\end{figure}

In the first simulation the environment is predefined deterministically. Figure \ref{fig: sim1 scene properties} shows the environment including the robot's trajectory, objects' locations and classes. Classes are represented by numbers, but for illustrations they are replaced by class types. The unsafe area of each class differs. Despite the fact that the environment is chosen deterministically, it is unknown to the robot.

Figure \ref{fig: sim1 expected reward} shows one run of $\psafe$ estimation versus time while the robot performs a pre-defined sequence of actions from Figure \ref{fig: sim1 scene properties}. For the \ebfull estimation, $ N^s = 10^6 $ samples were taken. Using the Hoeffding's inequality on the \ebfull estimation, it is guaranteed that the probability of error exceeding $0.3\%$ is less than $ 2.6 \cdot 10^{-14}$. Therefore, we compare the different estimators to this estimate since with high probability it is very close to the true $\psafe$. There were $10^3$  samples drawn for each of the remaining methods. Each estimator uses its own samples.
In Figure \ref{fig: sim1 expected reward}, \MCMC and \PFfull are aligned with the \ebfull, \SNIS The pruned methods result in biased results. There is no indication whether this bias will be positive or negative, and it can change from one time step to the next. Estimation with a positive bias can result in taking a risky action since $\psafe$ is assumed to be higher than it is. In contrast, estimation with a negative bias can lead to the rejection of a good action, i.e.~in a conservative behavior. Since \ebpruned, \PFpruned and \GSMAP are biased, they suffer from the issue.

\begin{figure}[H]
	\centering
	\includegraphics[width=0.70\linewidth]{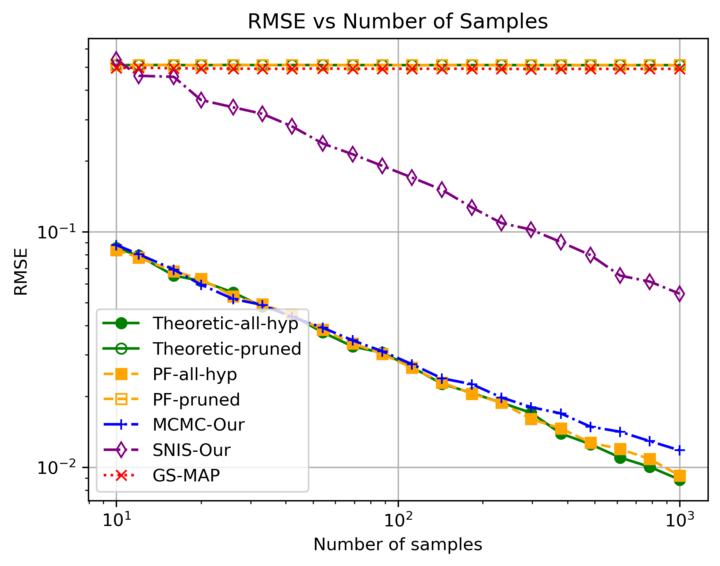}
	\caption{Expected reward error versus number of samples. The error decreases as the sample size increases for the \ebfull, \PFfull, \MCMC and \SNIS estimators. \Ebpruned, \PFpruned and \GSMAP estimators contain a bias that is not affected by sample size.}
	\label{fig: sim1 error vs samples}
\end{figure}

For the purpose of showing that the errors in the pruned estimators and in \GSMAP are caused by bias and not the sample size, we compared the root of the MSE (RMSE) versus the sample size in Figure \ref{fig: sim1 error vs samples}. We used the same simulation as in Figure \ref{fig: sim1 expected reward} previously. We calculated $\psafe$ at time-step 4 for an increasing sample size for each estimator. This process was repeated 100 times and the result was averaged. According to our results, the error decreases with an increase in sample size for \ebfull, \PFfull, \MCMC and \SNIS. For the pruned methods and \GSMAP, the error does not decrease with sample size, indicating a bias.

\begin{figure}[H]
	\includegraphics[width=0.70\linewidth]{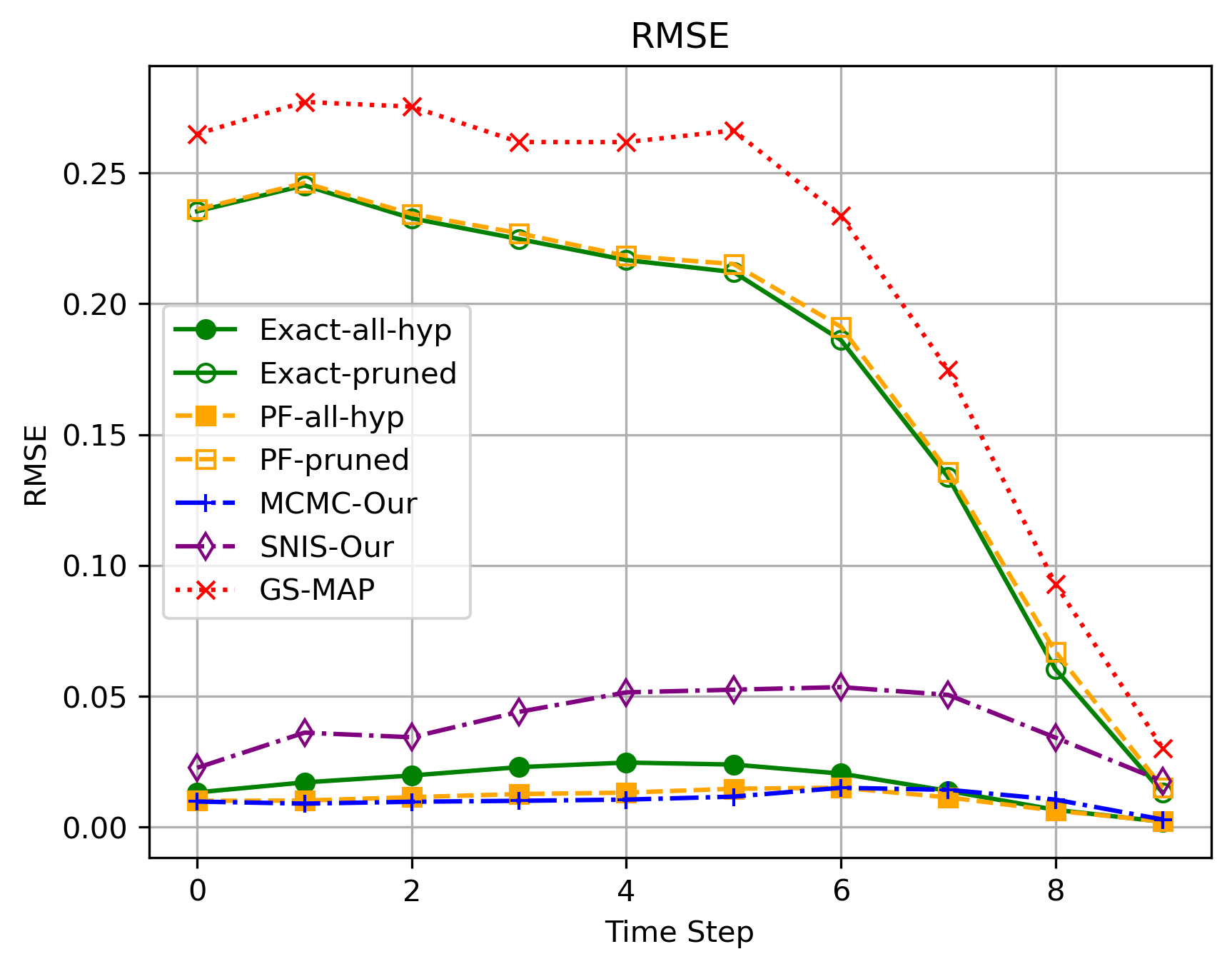}
	\centering
	\caption{RMSE of $\psafe$ estimations versus time-step.}
	\label{fig: sim1 rmse}
\end{figure}

In the following simulation objects' locations and classes were sampled randomly according to their prior probabilities. The robot's true trajectory is sampled from the transition model. At each time step, the robot receives geometric and semantic observations. In this simulation, $\psafe$ was estimated and the RMSE was evaluated, as depicted in Figure \ref{fig: sim1 rmse}. This simulation was repeated for $10^3$ trials. All methods use $10^3$ samples. The true expected value is computed using the \ebfull with $10^6$ samples. \MCMC, \PFfull and \ebfull achieve similar results. The RMSE of \SNIS is higher. However, the pruned versions and \GSMAP RMSEs are significantly higher.

\begin{figure}[H]
	\begin{subfigure}[b]{\linewidth}
		\centering
		\includegraphics[width=0.75\linewidth]{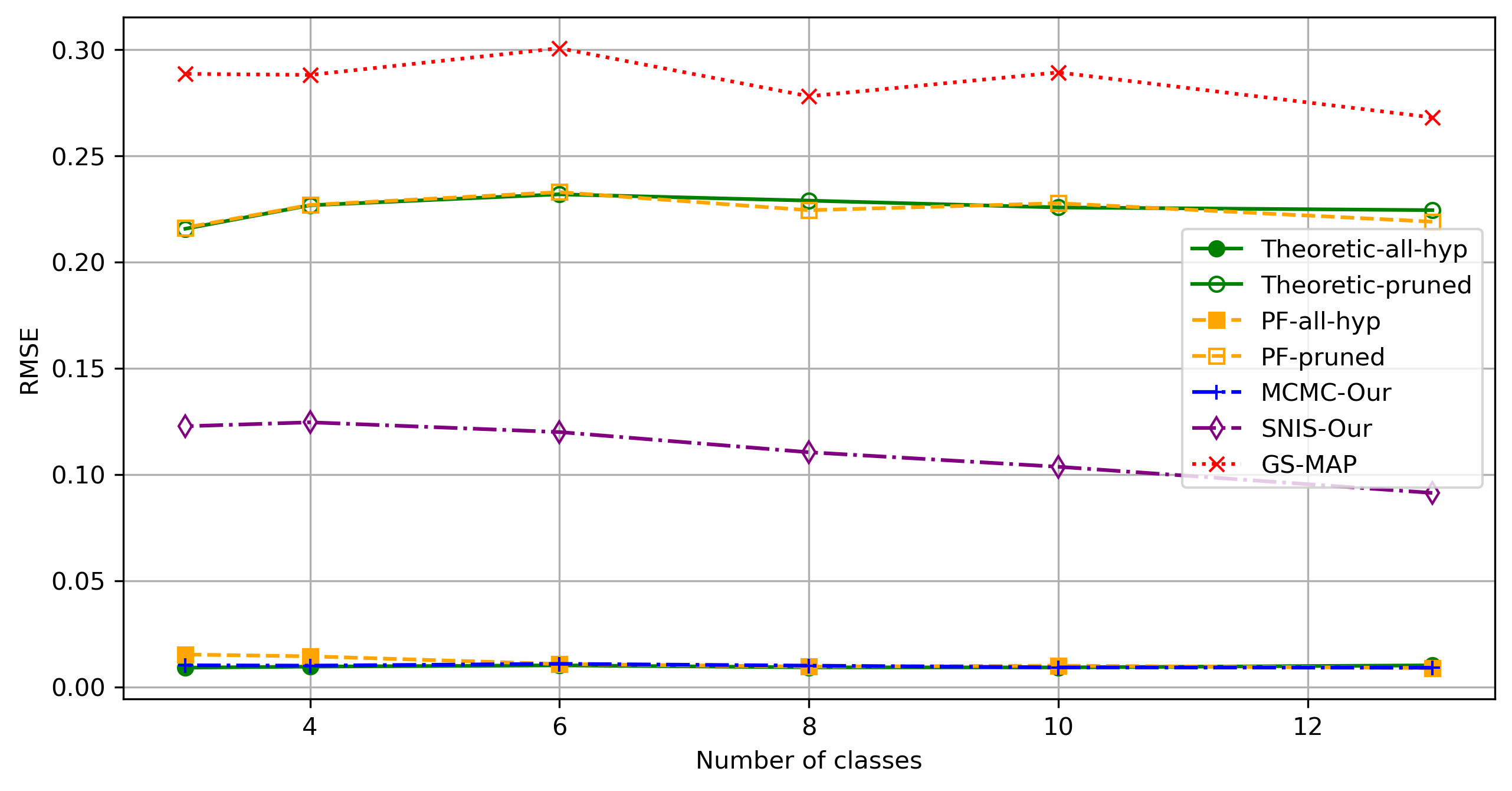}
		\caption{}
		\label{fig: rmse vs sample size}
	\end{subfigure}
	\begin{subfigure}[b]{\linewidth}
		\centering
		\includegraphics[width=0.75\linewidth]{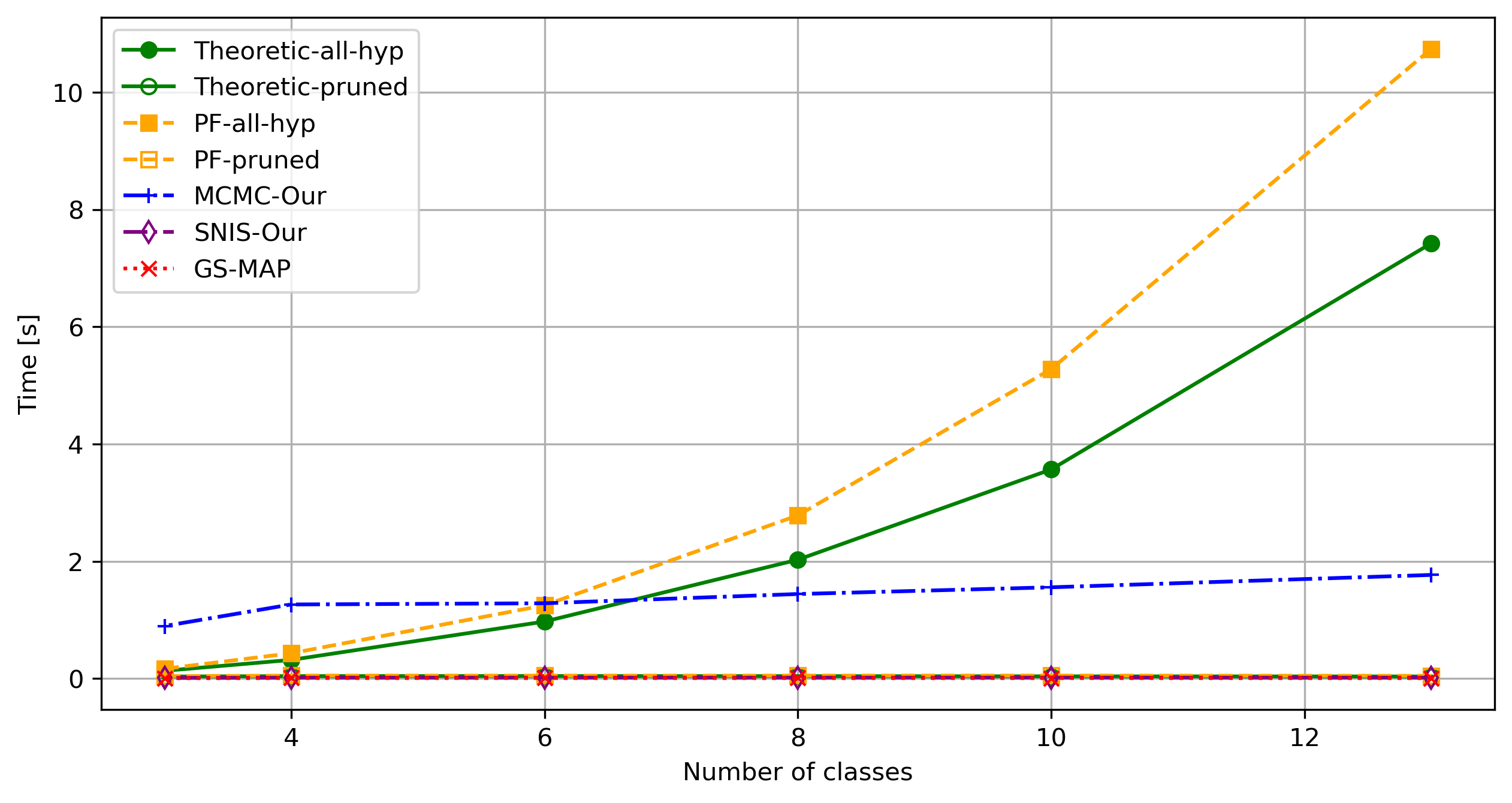}
		\caption{}
		\label{fig: time vs sample size}
	\end{subfigure}
	\caption{\label{fig:RunTime} \textbf{(a)} RMSE $\psafe$ estimations and \textbf{(b)} runtime versus number of classes. \MCMC achieves the same accuracy as the \ebfull and \PFfull estimates, but with a linear instead of exponential runtime in the number of classes. \SNIS has a slightly worse RMSE, but runs very efficiently. \Ebpruned, \PFpruned and \GSMAP  are significantly less accurate.}
\end{figure}

In Figure \ref{fig:RunTime}, 
the RMSE and running time are presented against the number of classes. The simulation consisted of 600 trials. The environment consists of 3 objects randomly selected according to the prior probability at the beginning of each trial. An increase in the number of classes increases the state space size, but does not change the state dimensionality. Results in Figure \ref{fig: rmse vs sample size} shows that the RMSEs of pruned estimates and \GSMAP estimate increase as the number of classes $N^c$ increases. Figure \ref{fig: time vs sample size} shows linear complexity of \MCMC, exponential complexity for \ebfull and \PFfull, and no noticeable increase in complexity for the pruned versions, GS-MAP, and \SNIS.

As expected, the runtime of the \ebfull and \PFfull increases exponentially with the number of classes, the runtime of the pruned methods, \SNIS and \GSMAP remains constant, and the runtime of \MCMC increases linearly.

\begin{figure}[H]
	\begin{subfigure}[b]{\linewidth}
		\includegraphics[width=0.75\linewidth]{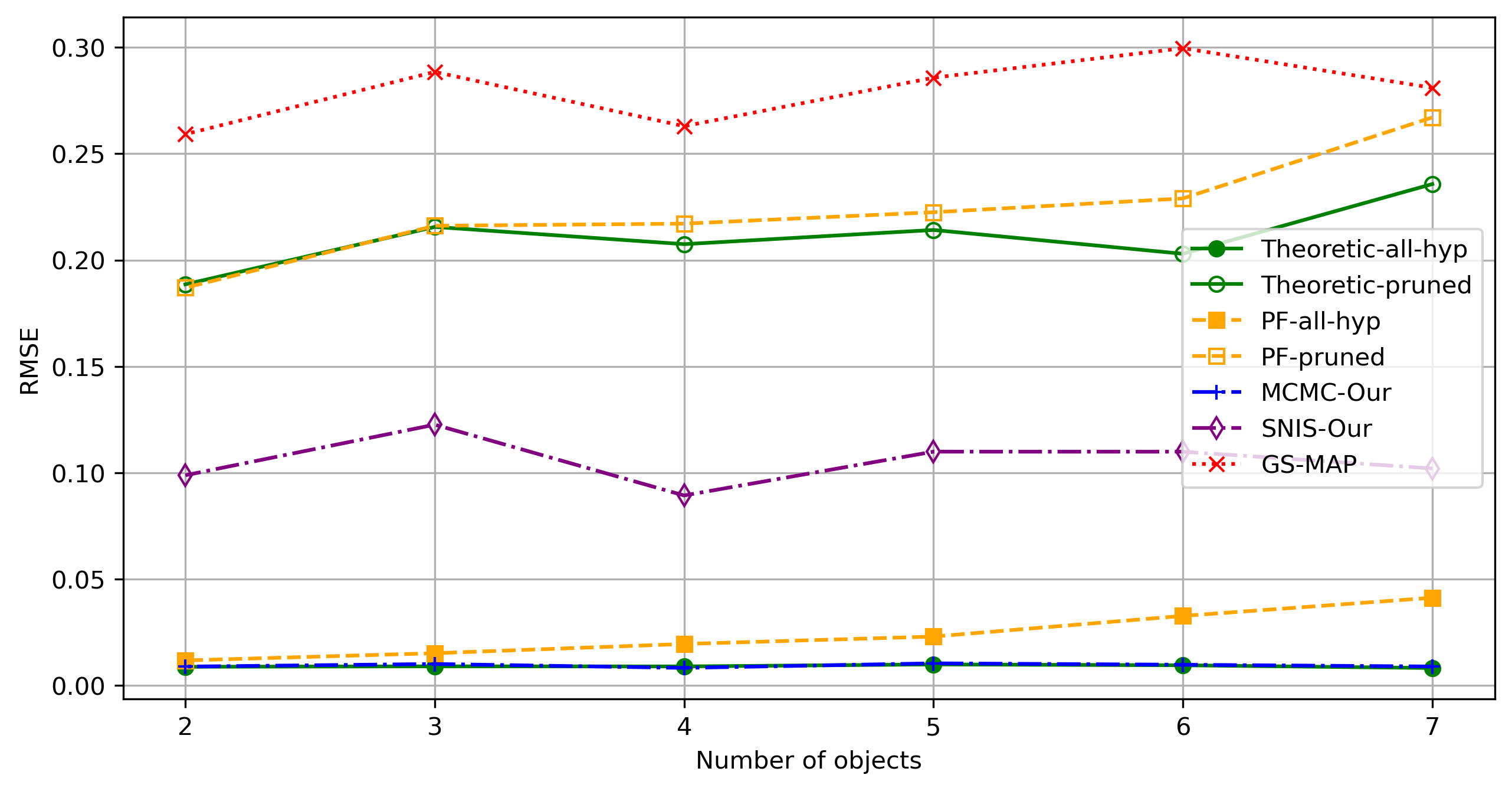}
		\centering
		\caption{}
		\label{fig: rmse vs number of objects}
	\end{subfigure}
	\begin{subfigure}[b]{\linewidth}
		\includegraphics[width=0.75\linewidth]{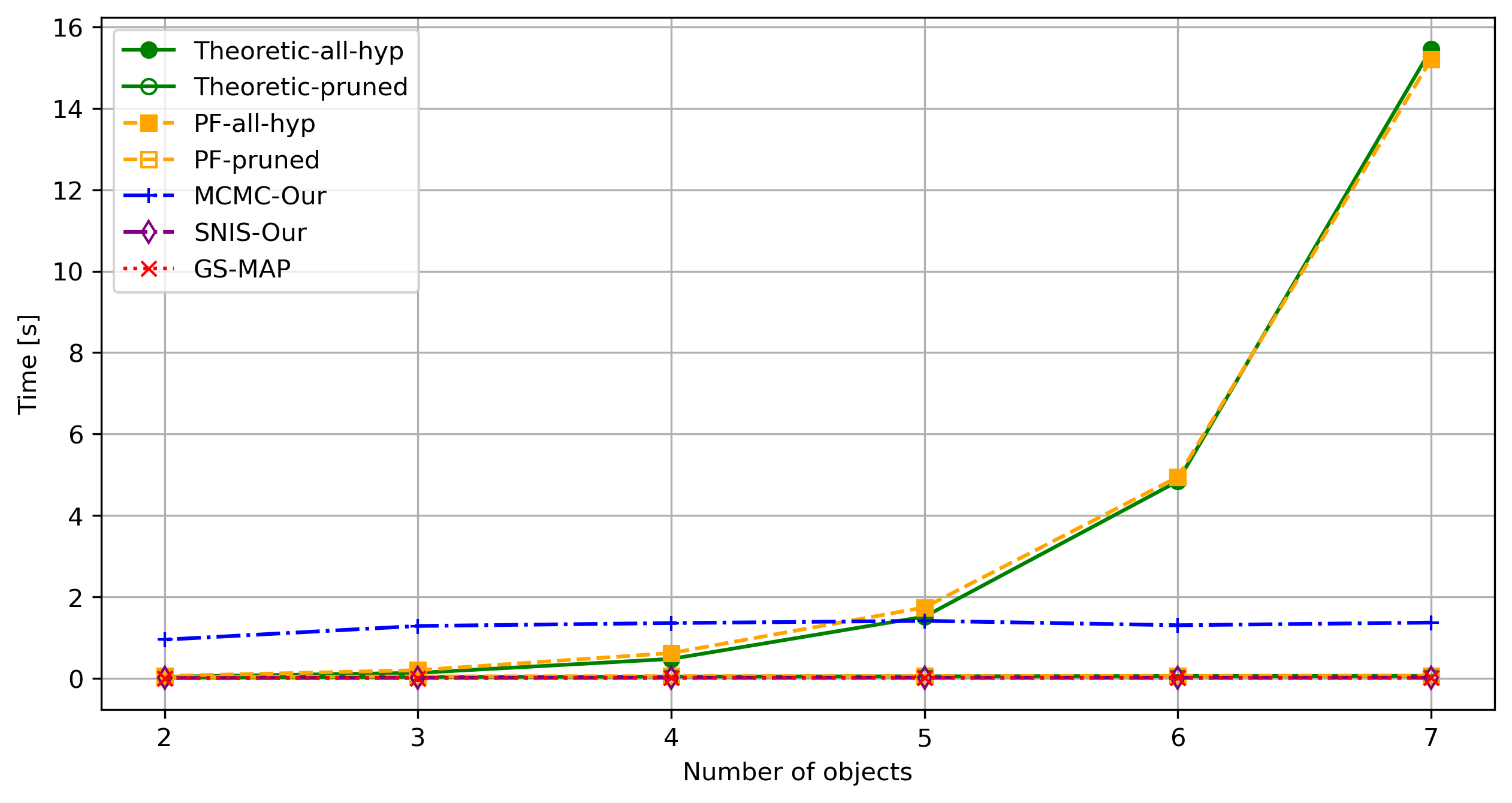}
		\centering
		\caption{}
		\label{fig: time vs number of objects}
	\end{subfigure}
	\caption{\label{fig:RMSE_RunTime} \textbf{(a)} RMSE of estimators of $\psafe$ and \textbf{(b)} running time versus number of objects.}
\end{figure}

In Figure \ref{fig:RMSE_RunTime}, 
the RMSE and running time are shown versus the number of objects. There are 600 trials conducted and the results are averaged. In all trials, the number of classes was 4. The environment consists of $N^o$ randomly selected objects at the beginning of each trial. An increase in the number of objects increases the dimensionality of the state. This affects the accuracy of the particle filter method. This is a well-known phenomenon in particle filtering known as the curse of dimensionality. Figure \ref{fig: rmse vs number of objects} shows an increase in RMSE as the number of objects increases for \PFfull. Consequently, the RMSE of \PFpruned estimator also increases. In contrast, the error in \ebfull, \MCMC, and \SNIS does not increase. As the number of objects increases, both \MCMC and \ebfull estimators maintain a very similar RMSEs.  

Figure \ref{fig: time vs number of objects} shows that the runtime of \ebfull and \PFfull estimators increase exponentially with the number of objects. In contrast, the runtime of pruned version, \SNIS and \GSMAP remains constant. \MCMC runtime increases linearly. This is consistent with the theoretical runtime of the estimators in Table \ref{table: methods performance}.

\subsection{Planning using the belief and safety constraint}\label{sec:Res_Planning}

\begin{figure}[h]
	\begin{subfigure}[b]{0.48\linewidth}
		\includegraphics[width=\linewidth]{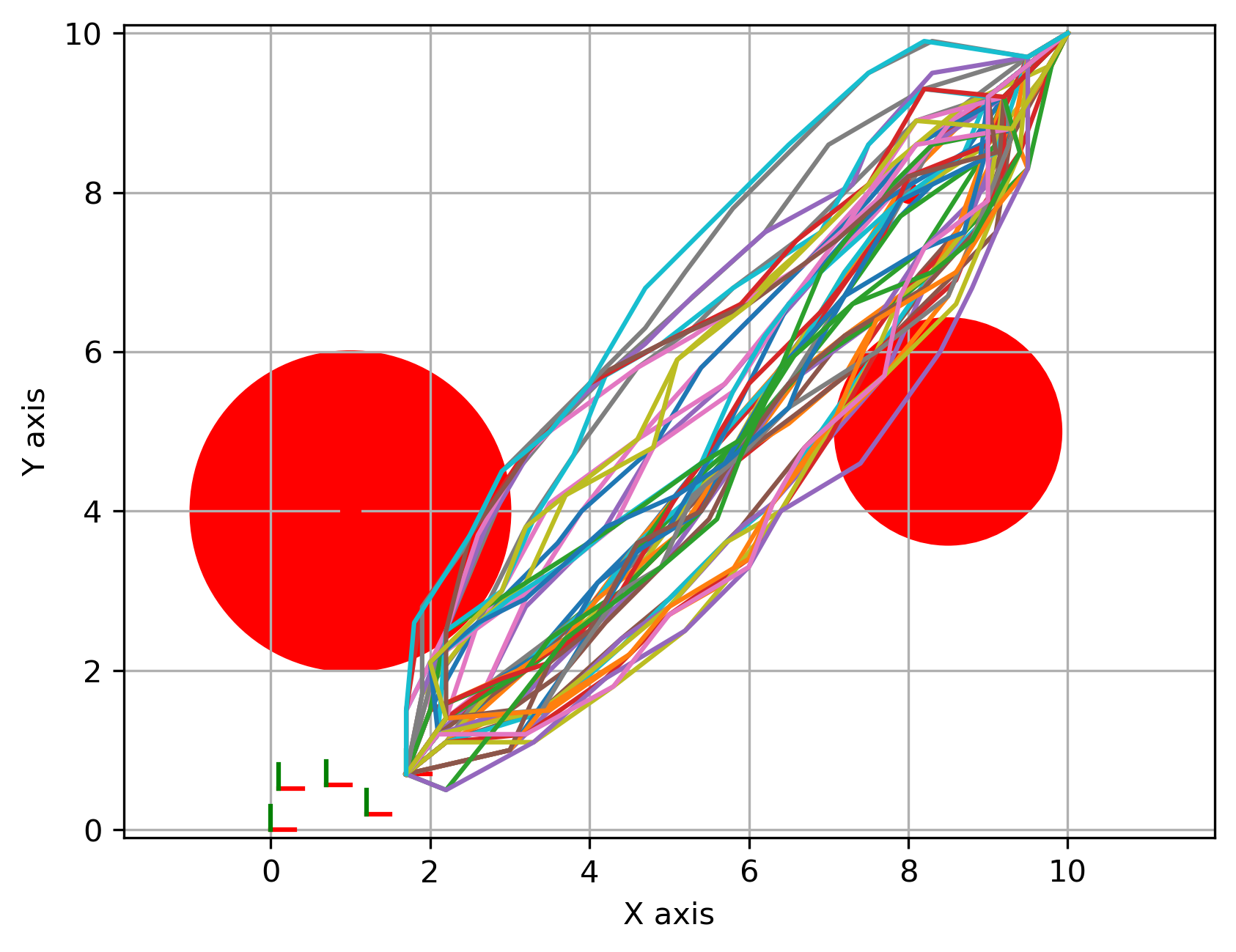}
		\centering
		\caption{}
		\label{fig: prm scene properties}
	\end{subfigure}
	\begin{subfigure}[b]{0.48\linewidth}
	\includegraphics[width=\linewidth]{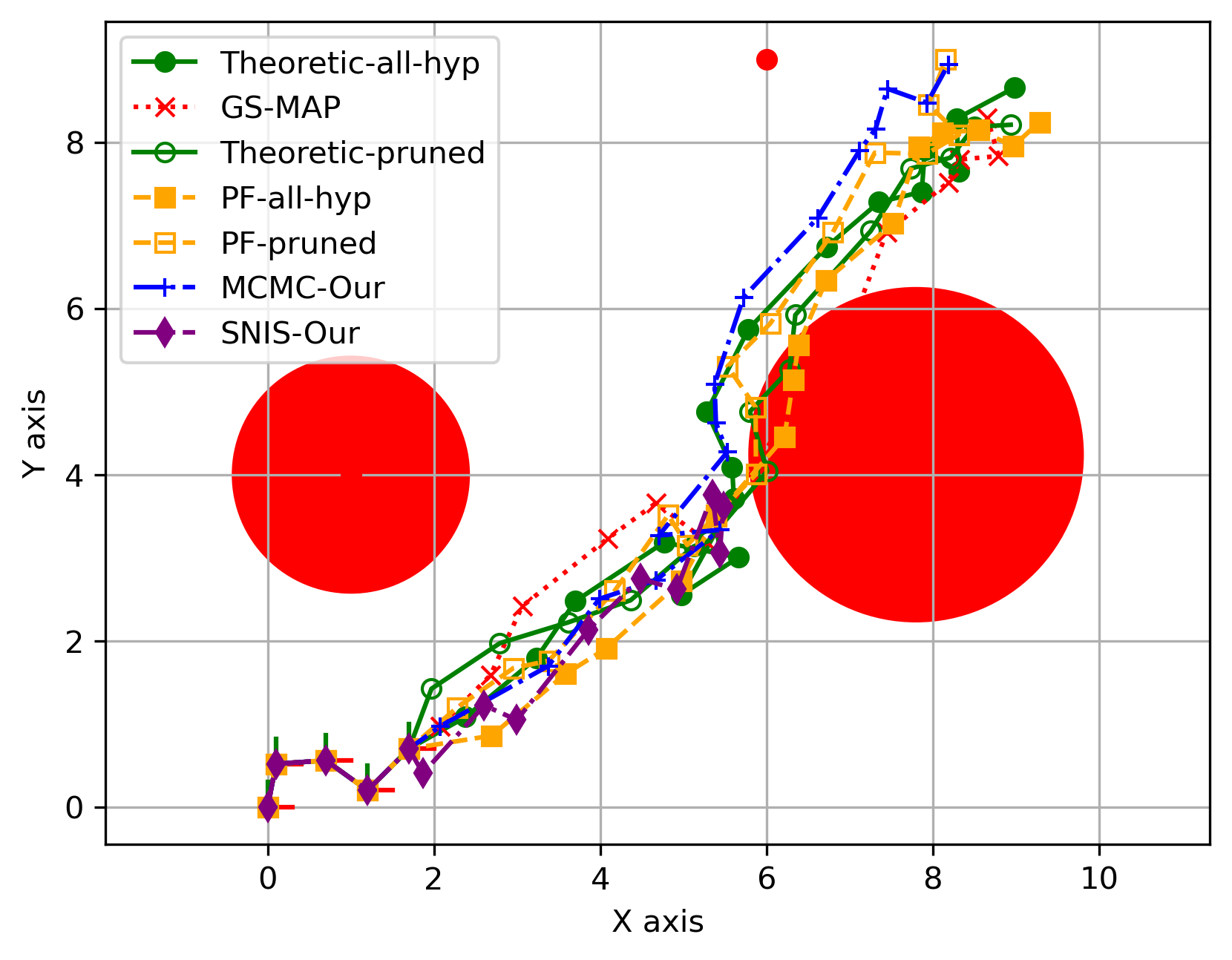}
	\centering
	\caption{}
	\label{fig: prm one trail}
	\end{subfigure}
	\caption{\textbf{(a)} Scene: four predefined actions, objects' \emph{true} unsafe areas in red, and $N$ shortest paths to goal from the PRM. \textbf{(b)} Chosen trajectory by each of the methods.}
\end{figure}

Here, we demonstrate the effect of different estimators on the safety constraint and the robot's actions. In this simulation, the robot decides what actions to take and executes them. The first four actions are the same for all methods and are predefined. Next, at each time-step the robot receives $N$ shortest paths from a probabilistic roadmap (PRM) and chooses the action that minimizes expected cost function under the constraint $\psafe\geq 0.95$. The cost function is defined as $r_t(\xx_t,a_t) = ||x_t - x_g|| + ||a_t||$ where $ x_g = (10,10)$ is the goal points. Each method estimates these terms differently, and therefore chooses a different path. Figure \ref{fig: prm scene properties} illustrates the true unsafe areas of the objects (in red) that correspond to the unknown true  classes of the objects, as well as the four predefined actions and the $N$ shortest paths from the PRM.

Upon receiving geometric and semantic observations and $N$ shortest paths, the robot calculates the probability of safety and expected reward for each path. The path that maximizes the expected reward and meets the safety constraints is selected. In the absence of a safe path the robot stops. In Figure \ref{fig: prm one trail}, we show one trial of the simulation.

In Table \ref{table: simulation_BSP_action_sequence}  the results of $100$ trials are summarized \TV{considering the number of hypotheses is $|\cc| = 16$ and $|\cc| = 10^9$}. 
\TV{In the case of $|\cc| = 16$,} the results indicate that \MCMC and \ebfull are the most accurate in evaluating $\psafe$ and the objective function, as they both achieved the highest success rate in finding a safe trajectory and minimizing the expected cost. 
They are also take the longest to run together with \PFfull. The running time of this simulation is consistent with those of previous simulations, Figures \ref{fig: time vs number of objects} and \ref{fig: time vs sample size}.
\TV{
In the case of the number of hypotheses $|\cc| = 10^9$, the results show that \MCMC and \SNIS maintain the highest accuracy where \SNIS bypasses \MCMC. In the case of \MCMC, this method produces samples that asymptotically converge to the theoretical belief, but as the theoretical belief becomes more complex, it becomes more difficult to converge to it. In terms of running time, these results are consistent with the previous findings, where all the methods take more time and the \MCMC take the longest time. There is also an increase in runtime with \ebpruned and \PFpruned as the number of classes increases from 4 to 1000.
}

\input{table_state_dependent_active_slam.tex}

%% file: table_state_dependent_active_slam.tex
\begin{table*}[h]
\centering
\begin{tabular}{ |p{3.05cm}||p{0.7cm}|p{1.5cm}|p{1.5cm}|p{1.5cm}||p{0.7cm}|p{1.5cm}|p{1.5cm}|p{2cm}| }
\hline
& \multicolumn{4}{c||}{\TV{$|\cc| = 16$}} & \multicolumn{4}{c|}{\TV{$|\cc| = 10^9$}} \\
\hline
Belief representation& $\psafe$ & Distance to goal if safe $\pm$ std&     
Trajectory length $\pm$ std  & \TV{Time[s] $\pm$ std} & \TV{$\psafe$} & \TV{Distance to goal if safe $\pm$ std} & \TV{Trajectory length $\pm$ std} & \TV{Time[s] $\pm$ std} \\
\hline
\hline
\Ebfull&        96/100& 3.74  $\pm$ 2.21&   14.75  $\pm$  2.46&  \TV{6.95 $\pm$ 4.14} & \TV{\NA}    & \TV{\NA}             & \TV{\NA}       & \TV{\NA}      \\
\Ebpruned&      91/100& 3.76  $\pm$ 2.52&   14.60  $\pm$  2.59&  \TV{4.01 $\pm$ 4.97} & \TV{75/100} & \TV{5.22 $\pm$ 4.93} & \TV{13.32 $\pm$ 4.67} & \TV{89.73 $\pm$ 104.81}\\
\PFfull&        87/100& 10.28 $\pm$ 4.54&   9.27   $\pm$  4.39&  \TV{8.99 $\pm$ 5.67} & \TV{\NA}    & \TV{\NA}             & \TV{\NA}       & \TV{\NA}      \\
\PFpruned&      81/100& 8.65  $\pm$ 4.62&   10.56  $\pm$  4.36&  \TV{4.06 $\pm$ 7.34} & \TV{80/100} & \TV{4.98 $\pm$ 4.22} & \TV{13.62 $\pm$ 4.09} & \TV{88.25 $\pm$ 106.71}\\
\TV{\GSMAP}&    \TV{68/100}& \TV{6.71  $\pm$ 5.29}&   \TV{12.03  $\pm$  5.19}&  \TV{2.17 $\pm$ 2.28} & \TV{58/100} & \TV{3.63 $\pm$ 3.41} & \TV{14.29 $\pm$ 3.12} & \TV{40.02 $\pm$ 42.68}\\
\textbf{\MCMC}& 96/100& 3.73  $\pm$ 2.31&   15.11  $\pm$  2.63&  \TV{4.49 $\pm$ 4.05} & \TV{83/100} & \TV{4.70 $\pm$ 4.46} & \TV{13.43 $\pm$ 4.11} & \TV{160.64 $\pm$ 149.33}\\
\textbf{\SNIS}& 94/100& 6.28  $\pm$ 3.02&   13.05  $\pm$  3.75&  \TV{3.97 $\pm$ 4.30} & \TV{88/100} & \TV{4.55 $\pm$ 3.85} & \TV{13.63 $\pm$ 3.52} & \TV{131.90 $\pm$ 156.43}\\
\hline
\end{tabular}
\caption{
Trials of executing optimal action under the safety constraint $\psafe \geq 0.95$. 
\TV{\NA indicates the method is not applicable due to runtime limitations. \Ebfull and \PFfull explicitly enumerate all hypotheses. Assuming constant runtime per hypothesis, evaluating a single action sequence would take approximately 1.4 months for \Ebfull and 1.9 months for \PFfull.}
}
\label{table: simulation_BSP_action_sequence}
\end{table*}

%% file: conclusions.tex
\section{Conclusions}
\label{Sec:conclusions}

We present a novel approach for estimating the value and objective function\TV{s in} hybrid semantic-geometric POMDP\TV{s}, where observation models and prior probabilities \TV{couple} the geometric and semantic variables. \TV{This coupling} causes all semantic variables to be dependent, \TV{resulting in} a combinatorial number of semantic hypotheses. \TV{In this setting, obtaining} representative samples of the belief required for estimating the value function and safety probability \TV{is highly challenging}. Under \TV{certain} assumptions, the MSE of sample-based estimators will grows exponentially with time. As a key contribution, we develop a novel factorization of the hybrid belief  and leverage it to provide two methods for obtaining representative samples that \TV{avoid} exponential MSE: (1) approximating samples \TV{from the theoretical} belief using MCMC, and (2) utilizing importance sampling with a proposal \TV{leveraging the belief factorization}.  As another key contribution, we show that under \TV{certain} assumptions and \TV{for specific} structured rewards, the expected reward can be computed efficiently with an explicit expectation over \emph{all} possible semantic mappings, further reducing estimation error. We prove that the probability of safety falls under this structured reward \TV{class}. Our empirical simulations show that our approaches achieve \TV{accuracy comparable to exhaustive enumeration of all semantic hypotheses, but with linear rather than exponential complexity}. We believe this work paves the way for more reliable and safer autonomous robots that operate in complex semantic-geometric environments.

%% file: appendix.tex
\appendix
\section{Appendix \label{sec: appendix}}

In the appendix we denote future observations $z_{k+1:t} $ as $z_k^+$, future policies $\pi_{k+1:T}$ as $\pi_k^+$, and future actions $a_{k:t-1}$ as $a_k^+ $.

\input{appendix_expected_reward_action_sequence.tex}

\input{appendix_proof_of_belief_formula.tex}

\input{appendix_sampling_illustration.tex}

\input{appendix_samples_lemma.tex}

\input{appendix_proof_of_expected_reward_brute_force_lemma.tex}

\input{appendix_proof_of_theorem_afficient_computation.tex}

\input{appendix_proof_of_probability_of_safety.tex}


%% file: appendix_expected_reward_action_sequence.tex
\subsection{Proof of Lemma \ref{lemma: expected reward action sequence}}
\label{sec: proof of lemma expected reward action sequence}

\begin{proof}
    \label{proof: expected reward action sequence}
The value function \eqref{eq:value function} is the sum of expected rewards.
We can examine a single expected reward since expectation is a linear
operation. A single expected reward at a future time $t$ is given by 
\TV{
\begin{multline}
\ee{z_{k:t}}\left[\rho_t\right]
=
\ee{z_{k:t}}\left[\ee{b_t\left[C,X_t\right]}\left[r_t\right]\right]
=\\
\intop_{z_{k:t}}\pr\left(z_{k:t}\mid H_k,\pi_{k:t}\right)
\sum_C\intop_{\xx_t}\pr\left(C,\xx_t\mid H_t\right)
r_t
.
\end{multline}
Using Bayes' theorem and chain rule,
we can extract the last observation $z_t$ from 
$\pr\left(C,\xx_t\mid H_t\right)$
and 
$\pr\left(z_{k:t}\mid H_k,\pi_{k:t}\right)$, respectively. Thus
\begin{multline}
\ee{z_{k:t}}\left[\rho_t\right]
=
\intop_{z_{k:t}}
\cancel{\pr\left(z_t\mid H_{t-1},a_{t-1}\right)}
\pr\left(z_{k:t-1}\mid H_k,\pi_{k:t-1}\right)
\times\\
\sum_C\intop_{\xx_t}
\frac{
        \PZ\left(z_t\mid C,\xx_t\right)
        \PT\left(x_t\mid a_{t-1},x_{t-1}\right)
        \pr\left(C,\xx_{t-1}\mid H_{t-1}\right)
    }
    {
        \cancel{\pr\left(z_t\mid H_{t-1},a_{t-1}\right)}
    }r_t
=\\
\intop_{z_{k:t-1}}
\pr\left(z_{k:t-1}\mid H_k,\pi_{k:t-1}\right)
\sum_C\intop_{\xx_t}
\pr\left(C,\xx_{t-1}\mid H_{t-1}\right)
\times\\
\intop_{x_t}
\PT\left(x_t\mid a_{t-1},x_{t-1}\right)
\intop_{z_t}
\PZ\left(z_t\mid C,\xx_t\right)
r_t
,
\label{eq:OneStepRecursiveExpectedReward}
\end{multline}
where $a_{t-1}$ is well defined since 
$a_{t-1} = \pi_{t-1}\left(b_{t-1}\right)$
and $b_{t-1}$ obtained given $H_{t-1}$. Recursively applying step
\eqref{eq:OneStepRecursiveExpectedReward} until time-step
$k$ yields
\begin{multline}
\ee{z_{k:t}}\left[\rho_t\right]
=
\sum_C\intop_{\xx_k}\pr\left(C,\xx_k\mid H_k\right)
\times\\
\intop_{x_{k+1}}\PT\left(x_{k+1}\mid a_k,x_k\right)
    \intop_{z_t}\PZ\left(z_{k+1}\mid C,\xx_{k+1}\right)
\cdots\\
\intop_{x_t}\PT\left(x_t\mid a_{t-1},x_{t-1}\right)
    \intop_{z_t}\PZ\left(z_t\mid C,\xx_t\right)r_t
\end{multline}
Finally, using expectation notation and considering that the value function is the sum of the expected rewards, the value function is obtained by}
    \begin{multline}
        V_k\left(b_k, \pi_{k:L} \right) 
        = \\
        \ee{b_k \left[C, \xx_k\right]}
        \prod_{\tau=k+1}^T
        \ee{\PT\left(x_{\tau} \mid \pi_{\tau-1},x_{\tau-1} \right)}
        \ee{\PZ\left(z_{\tau} \mid X^o,x_{\tau},C\right)}
        \left[
            \sum_{t=k}^{T}
            r_t
        \right]
        \label{eq: value function general reformulated in proof}
        .
    \end{multline}
    Assuming a state dependent reward $r_t = r_t\left(\xx_t, C, a_t\right)$ and open loop settings. Then, future observations do not participate in calculating the expectation in \eqref{eq: value function general reformulated in proof}. 
    As a result, future observations are marginalized
    \begin{multline}
        \label{eq: objective function state dep reward proof}
        J\left(b_k, a_{k:L} \right) 
        = \\
        \ee{b_k \left[C, \xx_k \right]}
        \prod_{\tau=k+1}^T
        \ee{\PT\left(x_{\tau} \mid a_{\tau-1},x_{\tau-1}\right)}
        \left[
            \sum_{t=k}^{T}
            r\left(C, \xx_t, a_t \right)
        \right].
    \end{multline}
\end{proof}

%% file: appendix_proof_of_belief_formula.tex
\subsection{Proof of Lemma \ref{lemma: belief formulation}}
\label{sec: proof of lemma belief formulation}


\begin{proof}
    \label{proof: belief formulation}
    Previously in \cite{Lemberg22iros}, we showed that the belief can be formulated as follows
    \begin{equation}
        b_k\left[\xx_k,C\right]
        = \\
        \frac{\tilde{\eta}_{1:k}}{\eta_{1:k}^g }
        b_k^g \left[ \xx_k \right] 
        \prod_{n=1}^{N^o}
        \belieftild{c_n \mid \xx_k }
        \label{eq: belief seperated},
    \end{equation}
    where $\eta_{1:k}^g$ is the normalization factor of the geometric belief \eqref{eq:geometricbel}.
    The following terms are defined for clarity.
    $\eta_{1:k} \define \frac{\tilde{\eta}_{1:k}}{\eta_{1:k}^g}$, 
    is the normalization factor of \eqref{eq: belief seperated},
    $
    \varphi_n\left(x_{1:k}, x^o_n \right)
    \define
    \sum_{c_n=1}^{N^c} \belieftild{c_n \mid \xx_k }
    $ 
    is the normalizer of $\belieftild{c_n \mid \xx_k }$,
    $ 
    \belief{c_n \mid \xx_k }
    = 
    \frac{\belieftild{c_n \mid \xx_k }}
        {\varphi_n\left(x_{1:k}, x^o_n \right)}
    $, 
    is the normalized semantic conditional belief of the $n$th object, 
    and
    $
    \Phi\left(\xx_k\right)
    \define
    \prod_{n=1}^{N^o}\varphi_n\left(x_{1:k}, x^o_n \right)
    $ is the total semantic contribution to the belief. 
    Accordingly, we can rewrite the belief as follows
    \begin{equation}
        \belief{C,\xx_k}
        =
        \eta_{1:k}
        \geobelief{\xx_k}
        \Phi\left(\xx_k\right)
        \prod_{n=1}^{N^o}
        \belief{c_n \mid \xx_k}
        \label{eq: belief X and C|X proof final}.
     \end{equation}
     Since the only term in \eqref{eq: belief X and C|X proof final} that is  dependent on the class of the $n$th object is $\belief{c_n \mid \xx_k}$, the marginal beliefs $\belief{c_n }$ and $\belief{\xx_k}$ are given by      
     \begin{equation}
        \belief{\xx_k}
        =
        \eta_{1:k}
        \geobelief{\xx_k}
        \Phi\left(\xx_k\right)
        ,
    \end{equation}
    \begin{equation}
        \belief{c_n}
        =
        \intop_{\xx_k}
        \belief{\xx_k}
        \belief{c_n \mid \xx_k}
        d\xx_k
        .
    \end{equation}

    \end{proof}

%% file: appendix_sampling_illustration.tex

\subsection{Proof of Theorem \ref{theorem:exponential MSE of IS}}
\label{sec: appendix important sampling illustration}

 We use explicit notation of the posterior probability $\pr(C\mid H_k)$ instead of $\belief{C}$,  since we are looking at the asymptotic behavior and considering different realizations of the true hypothesis $C^{tr}$ and history $H_k$.

We now proceed to the proof of Theorem \ref{theorem:exponential MSE of IS}.

\begin{proof}
\label{proof:exponential MSE of IS}    
Consider obtaining a realization of $C^{tr}$ and $H_k$, and consider using $N^s$
samples. Then, $\hat{\E}_{IS}^{\left(\xx,C\right)}$ is given by 
\begin{equation}
\hat{\E}_{IS}^{\left(\xx,C\right)}
\define
\frac{1}{N^s}
\frac{\pr(\xx_k^{(i)},C^{(i)} \mid H_k)}{q\left(\xx_k^{(i)},C^{(i)}\right)}
 g\left(\xx_k^{(i)},C^{(i)}\right)
.
\end{equation}
According to Theorem \ref{theorem: Rao-Blackwell for samples}, by calculating the explicit expectation
over $\xx_k$ and taking only samples of $C$, estimation MSE will
be reduced 
\begin{equation}
\mse\left(\hat{\E}^{IS\left(\xx,C\right)}\right)
\geq
\mse\left(\hat{\E}^{IS(C)}\right)
.
\label{eq:IS of C smaller then IS of C and X}
\end{equation}
Define 
$
\bar{g}\left(C\right)
=
\ee{\xx_k\mid C}\left[g\left(\xx_k,C\right)\mid C\right]
$,
thus 
$
\hat{\E}^{IS(C)}
\define
\frac{1}{N^s}
\frac{\pr(C^{(i)} \mid H_k)}{q\left(C^{(i)}\right)}\bar{g}(C^{(i)})
.
$
According to \cite{tokdar10sda} the MSE of $\hat{\E}^{IS(C)}$
is obtained by
\begin{multline*}
\mse\left(\hat{\E}^{IS(C)}\mid C^{tr}, H_k\right) 
=
\frac{
\E_{q}\left[\left(\frac{\pr(C \mid H_k)}{q(C)}\right)^2 \TV{\bar{g}}^2(C)\right]
}{N^s}
-
\mu_g^2
\\=
\sum_{C\in\cc}\frac{\pr^2(C\mid H_k)}{N^s q(C)}\bar{g}^2(C) 
\TV{
-
\mu_g^2
}
\end{multline*}
where $\mu_g$ is the expected value of $g\left(\xx_k,C\right)$.
Marginalizing over $H_k$ and $C^{tr}$, the MSE obtained by 
\begin{multline}
\mse\left(\hat{\E}^{IS}\right)
=
\ee{\pr_0(C^{tr})}
\ee{H_k\mid C^{tr}}
\left[
   \sum_{C\in\cc}\frac{\pr^2(C \mid H_k)}{N^s q(C)}\TV{\bar{g}}^2(C)
\right]
-\mu_g^2
.
\label{eq:mse exp IS}
\end{multline}
Moving $N^s$ and $\mu_g$ to the MSE side of \eqref{eq:mse exp IS} and
using \TV{Assumption} \ref{assumption:consistent}, one obtains
\small
\TV{
\begin{multline}
\left(\mse\left(
   \hat{\E}^{IS}\left[\bar{g}\right]
\right)+\mu_g^2\right)N^s 
= \\
\ee{\pr_0(C^{tr})}
\ee{H_k\mid C^{tr}}
\left[
   \left(
      \frac{\pr^2(C^{tr} \mid H_k)}{q(C^{tr})}
      +
      \sum_{C\in\cc\setminus \{C^{tr}\}}
      \frac{\pr^2(C \mid H_k)}{q(C)}
   \right)
   \bar{g}^{2} (C)
\right]
\\ \geq
\ee{\pr_0(C^{tr})}
\ee{H_k\mid C^{tr}}
\left[
   \frac{(1-\epsilon)^2}{q(C^{tr})}
   \TV{\bar{g}^{2} (C^{tr})}
\right]
\define 
L_2.
\label{eq:L1}
\end{multline}
}
\normalsize
Since $L_2$'s integrand does not include $H_k$, and $q(C)$ is assumed to be independent of $H_k$ (Assumption \ref{assumption:marginal}), $H_k$ can be marginalized from \eqref{eq:L1} 
\TV{
\begin{multline}
L_2
=
\ee{\pr_0(C^{tr})}
\left[
   \frac{(1-\epsilon)^2}{q(C^{tr})}
   \bar{g}^{2} (C^{tr})
\right]
\\=
(1-\epsilon)^2
\sum_{C^{tr}}
\frac{\pr_0(C^{tr})}{q(C^{tr})}
\bar{g}^2 (C^{tr})
.
\label{eq:bound mse}
\end{multline}
}
Define 
$
\tilde{w}(C)
=
\frac{\pr_0(C^{tr})}{q(C^{tr})}
,\;\;
\eta_w^{-1}
=
\sum_{C\in\cc}\TV{\tilde{w}}(C)$,
and 
$
w(C)
=
\eta_w \tilde{w}(C)
$.
$w(C)$ can be interpreted as a probability function.
Returning $\mu_g$ and $N^s$ to the right side of \eqref{eq:bound mse}
the lower bound of the MSE, denoted by $\underbar{MSE}$, is obtained by
\begin{align*}
\underbar{MSE} 
& =
\frac{1}{N^s}
(1-\epsilon)^2\eta_w^{-1}\sum_{C^{tr}}w(C)\TV{\bar{g}}^2 (C^{tr})-\mu_g^2\\
& =
\frac{1}{N^s}
(1-\epsilon)^2\eta_w^{-1}\underset{w}{\var}\left(\bar{g}\right)
\\& \approx
\frac{1}{N^s}
\eta_w^{-1}\underset{w}{\var}\left(\bar{g}\right)
.
\end{align*}
\TV{
Therefore, 
$
\lim_{k \to \infty} \underbar{MSE} 
= 
\frac{1}{N^s}
\eta_w^{-1}\underset{w}{\var}\left(\bar{g}\right)
$.
}
$\eta_w^{-1}$ is minimized when $q(C)=\pr_0(C)\;\;\forall C\in\cc$. 
This can be proven using Lagrange multipliers.
Therefore, the bound is obtained by 
\begin{equation}
\TV{\lim_{k \to \infty}}
\mse\left(\hat{\E}^{IS(C)}\right)
\geq
\frac{1}{N^s}
\left|\cc\right|\underset{w}{\var}\left(\bar{g}\right).
\end{equation}
Finally, according to \eqref{eq:IS of C smaller then IS of C and X}, \TV{one obtains}
\begin{align}
\TV{\lim_{k \to \infty}}
   \mse\left(\hat{\E}^{IS\left(\xx,C\right)}\right) \geq
\frac{1}{N^s}
\left|\cc\right|\underset{w}{\var}\left(\bar{g}\right).
\end{align}
\end{proof}

%% file: appendix_samples_lemma.tex
\subsection{Proof of Theorem \ref{theorem: Rao-Blackwell for samples}}

    
\begin{proof}
    \label{proof: Rao-Blackwell for samples}
    Consider approximating the expectation 
     $\ee{X,Y} \left[ g \left( X,C \right) \right]$
     using iid samples $X^{(i)},C^{(i)} \sim \pr\left[X,C\right], \:\: i=1,\cdots,N^s$,
     thus 
     $
     \hat{\E}^{(X,C)}
     \define
     \frac{1}{N^s}\sum_{i=1}^{N^s}g \left( X^{(i)},C^{(i)} \right).
     $
     This is an unbiased estimation with variance 
     $
     \var \left( \hat{\E}^{(X,C)} \right)
     =
     \frac{1}{N^s}\var \left( g \left( X,C \right) \right).
     $
     Consider another estimator by taking samples of $X$ and explicitly calculating the conditional expectation over $C$,  
     thus
     $
     \hat{\E}^{(X)}
     =
     \frac{1}{N^s}\sum_{i=1}^{N^s}
         \ee{C \mid X} \left[ g \left( X^{(i)},C \right) \mid X=X^{(i)} \right].
     $
     This is also an unbiased estimator with the variance 
     $
     \var \left( \hat{\E}^{(X)} \right)
     =
     \frac{1}{N^s} 
     \var \left( 
         \ee{C \mid X} \left[ g \left( X,C \right) \mid X \right] 
     \right)
     $.
     The difference between the variances is 
     \begin{multline}
         \var \left( \hat{\E}^{(X,C)} \right)
         -
         \var \left( \hat{\E}^{(X)} \right) 
         =\\
         =\frac{1}{N^s}\ee{X,C} 
         \left[ g^2 \left( X,C \right) \right]
         -
         \frac{1}{N^s}\ee{X} \left[ \ee{C \mid X}^2 \left[ g \left( X,C \right) \mid X \right] \right]\\
         =\frac{1}{N^s} \left( \ee{X} \left[ \ee{C \mid X} \left[ g^2 \left( X,C \right)-\ee{C \mid X}^2 \left[ g \left( X,C \right) \right] \mid X \right] \right] \right)\\
         =\frac{1}{N^s}\ee{X} \left[ \underset{C \sim \pr \left( C \mid X \right)}{\var} \left( g \left( X,C \right) \right) \right],
     \end{multline}
     where the last line is holds by using the identity 
     $\var \left( \psi \right)=\E \left[ \psi^2 \right]-\E^2 \left[ \psi \right]$
     inside the expectation on $X$. Since the variance of an unbiased estimator is the MSE, the theorem is proven.
\end{proof}

%% file: appendix_proof_of_expected_reward_brute_force_lemma.tex
\subsection{Proof of Lemma \ref{lemma: expected reward brute force}}

\begin{proof}
\label{proof: expected reward brute force}
\TV{
Consider a single expected reward component 
$\ee{z_k^+}
\left[
    \rho_t\left(b_t, a_t\right)
\right]
$
of the value function \eqref{eq:value function}. 
Formulating the expected reward explicitly and using Bayes' theorem, it can be derived as follows
\small
\begin{multline}
	\ee{z_k^+}
	\left[
		\rho_t\left(b_t, a_t\right)
	\right]
	=
	\intop_{z_k^+}
	\pr\left(z_k^+\mid H_k,\pi_k^+\right)
	\intop_{\xx_t}\sum_C
	\pr\left(\xx_t,C\mid H_t\right)r_t
	=\\
	\intop_{z_k^+}
	\pr\left(z_k^+\mid H_k,\pi_k^+\right)
	\intop_{\xx_t}\sum_C
	\pr\left(\xx_t,C\mid H_k,z_k^+,\pi_k^+\right)
	r_t
	=\\
	\intop_{z_k^+}\cancel{\pr\left(z_k^+\mid H_k,\pi_k^+\right)}\intop_{\xx_t}\sum_C\frac{\pr\left(z_k^+\mid\xx_t,C\right)\pr\left(\xx_t,C\mid H_k,\pi_k^+\right)}{\cancel{\pr\left(z_k^+\mid H_k,\pi_k^+\right)}}r_t
	=\\
	\intop_{z_k^+}\intop_{\xx_t}\sum_C\pr\left(z_k^+\mid\xx_t,C\right)\pr\left(\xx_t,C\mid H_k,\pi_k^+\right)r_t
	,
	\label{eq:ProofExpectedReward1}
\end{multline}
\normalsize
where the future beliefs $\belief[t]{\xx_t, C}$ and actions $a_t$ are defined uniquely by $H_k$, the future observations $z_k^+$, and the policies $\pi_k^+$. The future beliefs and action can be calculated as by defining the update operator $b_{\tau+1} \define \varphi(b_{\tau}, a_{\tau}, z_{\tau+1})$. The they are obtained recursively by $a_{\tau} = \pi_{\tau}\left(b_{\tau}\right)$ and $b_{\tau+1} = \varphi\left(b_{\tau}, a_{\tau}, z_{\tau+1}\right)$. 
Using chain rule we can refactor the \TV{following} probability
\begin{equation}
	\pr\left(\xx_t,C\mid H_k,\pi_k^+\right)=\pr\left(\xx_t\mid H_k,\pi_k^+\right)\pr\left(C\mid\xx_t,H_k,\pi_k^+\right)
\end{equation}
The semantic hypothesis $C$ is dependent on $\xx_t$ only through the prior probability and the observations that connect between them, therefore, 
\begin{equation}
	\pr\left(C\mid\xx_t,H_k,\pi_k^+\right)=\pr\left(C\mid\xx_k,H_k\right)
	.
	\label{eq:UsingChainRule}
\end{equation}
By substituting \eqref{eq:UsingChainRule} into \eqref{eq:ProofExpectedReward1}, we obtain
\begin{multline}
	\ee{z_k^+}\left[\rho_t\left(b_t\right)\right]
	=\\
	\intop_{z_k^+}\intop_{\xx_t}\sum_C\pr\left(z_k^+\mid\xx_t,C\right)\pr\left(\xx_t\mid H_k,\pi_k^+\right)\pr\left(C\mid\xx_k,H_k\right)r_t
	.
\end{multline}
This can be reformulated as follows
\begin{equation}
	\ee{z_k^+}\left[\rho_t\left(b_t, a_t\right)\right]
	=
	\ee{\pr\left(\xx_t\mid H_k,\pi_k^+\right)}
	\left[
		\ee{\belief{C\mid\xx_k}}
		\prod_{\tau=k+1}^t
		\ee{\PZ\left(z_{\tau}\mid\xx_{\tau},C\right)}
		\left[r_t(\cdot)\right]
	\right]
	.
\end{equation}
Using the assumption that $r_t$ is state dependent, and it is not dependent on the action, or it is dependent on the action but the actions are predetermined in an open loop settings, then it is possible to marginalize the observations.
\begin{equation}
	\ee{z_k^+}\left[\rho_t\left(b_t, a_t\right)\right]
	=
	\ee{\pr\left(\xx_t\mid H_k,\pi_k^+\right)}
	\left[
		\ee{\belief{C\mid\xx_k}}
		\left[r_t(\cdot)\right]
	\right]
	.
	\label{eq:ExpectedRewardMarginalizedObservations}
\end{equation}
Finally, the value function is the sum of the expected rewards, therefore it can be written as follows
\begin{equation}
	V_k\left(b_k, \pi_{k:L} \right) 
	=
	\ee{\pr\left(\xx_L \mid H_k, \pi_{k:L-1}\right)}
	\left[
	\ee{\belief{C \mid \xx_k}}
	\left[
	\sum_{t=k}^{L}
	r_{t}\left(\cdot\right)
	\right]
	\right]
	.
\end{equation}
}
\end{proof}

%% file: appendix_proof_of_theorem_afficient_computation.tex
\subsection{Proof of Theorem~\ref{theorem: expected reward efficient}}
\label{sec: proof of theorem expected reward efficient}


\begin{proof}
\label{proof: expected reward state dependent}

Consider the reward in \eqref{eq: efficient reward}.
Define the following augmented element
\begin{equation}
\tilde{r}_{t,j,n}\left(c_n, \xx_t\right)
\define 
\begin{cases}
r_{t,j,n}\left(c_n, \xx_t\right) & n \in \theta_j\\
1 & \text{otherwise}
\end{cases}.
\label{eq: augmented reward def}
\end{equation}
\TV{Using the augmented element \eqref{eq: augmented reward def}, the reward \eqref{eq: efficient reward} can be rewritten as follows}
\begin{equation}
r_t \left(C,\xx_t\right)
=
\sum_{j=1}^{|\Theta|} 
\prod_{n=1}^{N^o} 
\tilde{r}_{t,j,n}\left(c_n, \xx_t\right).
\label{eq: augmented reward}
\end{equation}
Since expectation is a linear operation, we can take the sum in \eqref{eq: augmented reward} out of the expectation and return it latter.
For now, consider that the reward above \eqref{eq: augmented reward} \TV{consists} of only one element of the sum,
\begin{equation}
	r_t \left(C,\xx_t\right)=\prod_{n=1}^{\TV{N^o}}\tilde{r}_{t,j,n}(c_n, \xx_t)
	.
	\label{eq: efficient reward one element}	
\end{equation}
Consider the expected reward \eqref{eq: expected reward brute force} 
and using the above inner reward function \eqref{eq: efficient reward one element}, the expected reward can be formulated as follows
\begin{multline}
	\ee{z_{k+1:t}}\left[\rho\left(b_t\right)\right]
	=\\
	\ee{\pr\left(\xx_t \mid H_k, \pi_{k:t-1}\right)}
	\left[
	\ee{\belief{C \mid \xx_k}}
	\left[
		\prod_{n=1}^{\TV{N^o}}\tilde{r}_{t,j,n}(c_n, \xx_t)
	\right]
	\right]
	=\\
	\ee{\pr\left(\xx_t \mid H_k, \pi_{k:t-1}\right)}
	\sum_{\TV{C}}
	\left[
		\prod_{n=1}^{\TV{N^o}}
		\belief{c_n \mid \xx_k}
		\tilde{r}_{t,j,n}(c_n, \xx_t)
	\right]
	\label{eq:one element expected reward}.
\end{multline}
To simplify notations we will use 
$\pr^{+}(\xx_t) = \pr\left(\xx_t \mid H_k, \pi_{k:t-1}\right)$.
Since both $\belief{c_n \mid \xx_k}$ and $\tilde{r}_{t,j,n}(c_n, \xx_t)$ are dependent on a single object \TV{class $c_n$}, the expected reward \eqref{eq:one element expected reward} can be reorganized as follows
\begin{equation}
\ee{z_{k+1:t}}\left[\rho\left(b_t\right)\right]
=
\ee{\pr^{+}(\xx_t)}
\left[
	\prod_{n=1}^{N^o}\sum_{c_n=1}^{N^c}
	\belief{c_n \mid \xx_k}
	\tilde{r}_{t,j,n}\left(c_n,\xx_t\right)		
\right]
.
\label{eq:expected value state dependent tmp1}
\end{equation}
If $n \notin \theta_j$ the augmented element is equal to 1 and the sum in \eqref{eq:expected value state dependent tmp1} result in 
$\sum_{c_n=1}^{N^c} \belief{c_n \mid \xx_k} \cdot 1$. 
Therefore we can simplify \eqref{eq:expected value state dependent tmp1} to
\begin{equation}
	\ee{z_{k+1:t}}\left[\rho\left(b_t\right)\right]
	=
	\ee{\pr^{+}(\xx_t)}
	\left[
		\prod_{n \in \theta_j}
		\ee{\belief{c_n \mid \xx_k}}
		\left[
		r_{t,j,n}\left(c_n,\xx_t\right)
		\right]
	\right]
	.
	\label{eq: expected reward state dependent tmp2}
\end{equation}
By restoring the full reward \eqref{eq: efficient reward} 
into the expected reward \eqref{eq: expected reward state dependent tmp2}, one obtains 
\begin{equation}
	\ee{z_{k+1:t}}\left[\rho\left(b_t\right)\right]
	=
	\ee{\pr^{+}(\xx_t)}
	\left[
		\sum_{j=1}^{|\Theta|}
		\prod_{n \in \theta_j}
		\ee{\belief{c_n \mid \xx_k}}
		\left[
		r_{t,j,n}\left(c_n,\xx_t\right)
		\right]
	\right]
	.
	\label{eq: expected reward state dependent}
\end{equation}

The approximation of the expected reward, we will use samples of 
$\xx_t^{(i)} \sim \pr^{+}(\xx_t)$
and approximate it as follows
\begin{multline}
	\hat{\E}[\rho_t(b_t)]
	= \\
	\frac{1}{N^s}
	\sum_{i=1}^{N^s}
	\sum_{j=1}^{|\Theta|}
	\prod_{n\in \theta_j}
	\sum_{c_n=1}^{N^c}	
	\belief{c_n \mid \xx_k^{(i)}}
	r_{t,j,n}\left(c_n,\xx_t^{(i)}\right)
	.
\end{multline}
The computation complexity is $O\left(N^s \Thetasize N^c \right)$.

\end{proof}

%% file: appendix_proof_of_probability_of_safety.tex
\subsection{Proof of Proposition~\ref{prop: safety inner reward}}

\begin{proof}
\label{proof: safety inner reward}
Let $\rho$ be defined in \eqref{eq:reward}. $\psafe$ \eqref{eq:psafe} can be formulated as an expectation over $\xx_L, C \mid b_k, \pi_{k:L-1}$ on the indicator function 
$\oneb \left[\wedge_{\tau=k+1}^L x_{\tau}\notin \xunsafe \left(C, X^o\right) \right]$, 
thus
\begin{equation}
    \psafe
    =
    \E \left[
        \oneb \left[\wedge_{\tau=k+1}^L x_{\tau}\notin \xunsafe \left(C, X^o\right) \right]
        \mid b_k, \pi_{k:L}  
    \right]
    .
    \label{eq:psafe 1}
\end{equation}
Marginalizing over future observations, we obtain
\begin{multline}
    \psafe
    =\\
    \ee{z_{k:L}|b_k,\pi_{k:L}} 
    \left[
    \ee{\xx_L,C|b_L}
    \left\{
    \oneb \left[\wedge_{\tau=k+1}^L x_{\tau}\notin \xunsafe \left(C, X^o\right) \right] 
    \right\}
    \right]
    .
    \label{eq:psafe 2}
\end{multline}
This is a state-dependent reward, with the indicator being the inner reward function
$
r_{safe}\left(\xx_t, C\right) 
=
\oneb \left[\wedge_{\tau=k+1}^L x_{\tau}\notin \xunsafe \left(C, X^o\right) \right]
$.
Furthermore, $r_{safe}$ can be formulated as follows
\begin{multline}
    r_{safe}\left(\xx_t, C\right) 
    = 
    \oneb
    \left[
        \wedge_{\tau=k+1}^L x_{\tau}\notin \cup_{n=1}^{N^o} \xunsafe \left(c_n, x^o_n\right)
    \right]\\
    =
    \prod_{n=1}^{N^o}
    \prod_{\tau=k+1}^L
    \oneb \left[x_{\tau} \notin \xunsafe \left(c_n, x^o_n\right) \right]
    .
    \label{eq:safety condition}
\end{multline}

\end{proof}